\documentclass{article}

\usepackage[nohyperref,accepted]{icml2019}

\usepackage[utf8]{inputenc}
\usepackage[T1]{fontenc}
\usepackage{url}
\usepackage{amsfonts}
\usepackage{nicefrac}
\usepackage{microtype}
\usepackage{amsmath}
\usepackage{amssymb}
\usepackage{mathtools}
\usepackage[english]{babel}
\usepackage{subcaption}
\usepackage{ragged2e}
\usepackage{booktabs}
\usepackage{tikz}
\usepackage{etoolbox}
\usepackage{xspace}
\usepackage{xpatch}
\usepackage{xstring}
\usepackage{enumerate}
\usepackage[bookmarks=false]{hyperref}
\usepackage{natbib}
\usepackage{setspace}
\usepackage{tabularx}
\usepackage{wrapfig}
\usepackage{tikz}
\usepackage[capitalise,noabbrev,nameinlink]{cleveref}
\usepackage[useregional=numeric]{datetime2}
\usepackage[linesnumbered,ruled,noend]{algorithm2e}
\usepackage{graphicx}
\usepackage{enumitem}
\usepackage{tikzscale}
\usepackage{afterpage}
\usepackage{multicol}

\usetikzlibrary{positioning,arrows.meta,fit,calc,backgrounds,decorations.pathmorphing}

\definecolor{mydarkblue}{rgb}{0,0.08,0.45}
\hypersetup{%
colorlinks=true,
linkcolor=mydarkblue,
citecolor=mydarkblue,
filecolor=mydarkblue,
urlcolor=mydarkblue}

\newcommand{\paritem}[1]{\item\textbf{#1}\quad}
\setlist[itemize]{leftmargin=1.5em}  
\xapptocmd\normalsize{%
\abovedisplayskip=5pt plus 3pt minus 4pt
\abovedisplayshortskip=0pt plus 2pt
\belowdisplayskip=5pt plus 3pt minus 4pt
\belowdisplayshortskip=3pt plus 3pt minus 2pt
}{}{}

    \allowdisplaybreaks  
\newcommand{\eqbr}{\,\hookleftarrow\\[-1ex]}  
\DeclareRobustCommand{\note}[1]{}

\newcommand{\method}{PlaNet\xspace}
\newcommand{\fullmethod}{Deep Planning Network\xspace}

\setcitestyle{authoryear,round,citesep={;},aysep={,},yysep={;}}
\creflabelformat{equation}{#2\textup{#1}#3}
\crefname{algocf}{Algorithm}{Algorithms}
\Crefname{algocf}{Algorithm}{Algorithms}
\let\oldbibitem\bibitem
\def\bibitem{\vfill\oldbibitem}

\graphicspath{{figures/}}

\newcommand{\rot}[1]{\rotatebox{90}{\shortstack[l]{#1}}}

\renewcommand*\d{\mathop{}\!\textnormal{\slshape d}}
\newcommand*\I{\mathop{}\!\mathbb{I}}

\newcommand{\describe}[2]{\underbracket[1pt][0pt]{#1}_\text{\makebox[1em][c]{#2}}}

\DeclarePairedDelimiter{\ceil}{\lceil}{\rceil}

\DeclarePairedDelimiterX{\SquareBrackets}[1]{[}{]}{#1}
\DeclarePairedDelimiterX{\RoundBrackets}[1]{(}{)}{#1}
\DeclarePairedDelimiterX{\DivergenceBrackets}[2]{[}{]}{#1\;\delimsize\|\;#2}

\NewDocumentCommand{\pr}{ O{p} r() }{
  \def\prArg{#2}\patchcmd{\prArg}{|}{\mid}{}{}#1\RoundBrackets{\prArg}}
\NewDocumentCommand{\p}{ r() }{\pr[p](#1)}
\NewDocumentCommand{\q}{ r() }{\pr[q](#1)}
\NewDocumentCommand{\prm}{ r() }{\pr[\mathrm{p}](#1)}
\NewDocumentCommand{\Normal}{ r() }{\pr[\operatorname{Normal}](#1)}
\NewDocumentCommand{\Cat}{ r() }{\pr[\operatorname{Cat}](#1)}
\NewDocumentCommand{\Beta}{ r() }{\pr[\operatorname{Beta}](#1)}
\NewDocumentCommand{\Bernoulli}{ r() }{\pr[\operatorname{Bernoulli}](#1)}
\NewDocumentCommand{\Dir}{ r() }{\pr[\operatorname{Dir}](#1)}

\newcommand{\E}[3][]{\mathbb{\operatorname{E}}_{#2}#1[#3#1]}
\newcommand{\MI}[1]{\mathbb{\operatorname{I}}\RoundBrackets{#1}}
\renewcommand{\H}[1]{\mathbb{\operatorname{H}}\RoundBrackets{#1}}

\newcommand{\KL}{\mathrm{\operatorname{KL}}\DivergenceBrackets}
\newlength\widthE
\newcommand{\Ebelow}[3][]{\settowidth\widthE{$\operatorname{E}$}
\mathop{\operatorname{E}}_{\vphantom{|^|}\mathmakebox[0.5\widthE][l]{#2}}#1[#3#1]}

\icmltitlerunning{Learning Latent Dynamics for Planning from Pixels}

\begin{document}

\twocolumn[
\icmltitle{Learning Latent Dynamics for Planning from Pixels}
\begin{icmlauthorlist}
\icmlauthor{Danijar Hafner}{gb,ut}
\icmlauthor{Timothy Lillicrap}{dm}
\icmlauthor{Ian Fischer}{gr}
\icmlauthor{Ruben Villegas}{gb,um} \\
\icmlauthor{David Ha}{gb}
\icmlauthor{Honglak Lee}{gb}
\icmlauthor{James Davidson}{gb}
\end{icmlauthorlist}
\icmlaffiliation{gb}{Google Brain}
\icmlaffiliation{dm}{DeepMind}
\icmlaffiliation{gr}{Google Research}
\icmlaffiliation{ut}{University of Toronto}
\icmlaffiliation{um}{University of Michigan}
\icmlcorrespondingauthor{Danijar Hafner}{mail@danijar.com}
\icmlkeywords{latent planning, model-based reinforcement learning, variational inference, Bayesian filtering, model predictive control, video prediction}
\vskip 0.3in
]
\printAffiliationsAndNotice{}

\begin{abstract}
\begin{hyphenrules}{nohyphenation}
Planning has been very successful for control tasks with known environment dynamics. To leverage planning in unknown environments, the agent needs to learn the dynamics from interactions with the world. However, learning dynamics models that are accurate enough for planning has been a long-standing challenge, especially in image-based domains. We propose the \fullmethod (\method), a purely model-based agent that learns the environment dynamics from images and chooses actions through fast online planning in latent space. To achieve high performance, the dynamics model must accurately predict the rewards ahead for multiple time steps. We approach this using a latent dynamics model with both deterministic and stochastic transition components. Moreover, we propose a multi-step variational inference objective that we name latent overshooting. Using only pixel observations, our agent solves continuous control tasks with contact dynamics, partial observability, and sparse rewards, which exceed the difficulty of tasks that were previously solved by planning with learned models. \method uses substantially fewer episodes and reaches final performance close to and sometimes higher than strong model-free algorithms.
\end{hyphenrules}
\end{abstract}

\section{Introduction}
\label{introduction}

\afterpage{\begin{figure*}[t]
\providecommand{\width}{}
\renewcommand{\width}{0.15\textwidth}
\centering
\begin{subfigure}[t]{\width}
\includegraphics[width=\textwidth]{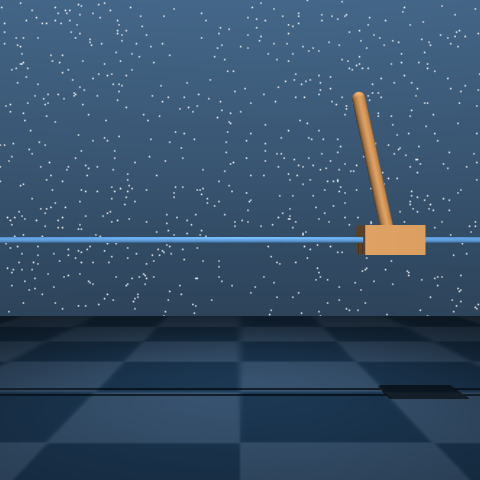}
\caption{Cartpole}
\end{subfigure}\hfill%
\begin{subfigure}[t]{\width}
\includegraphics[width=\textwidth]{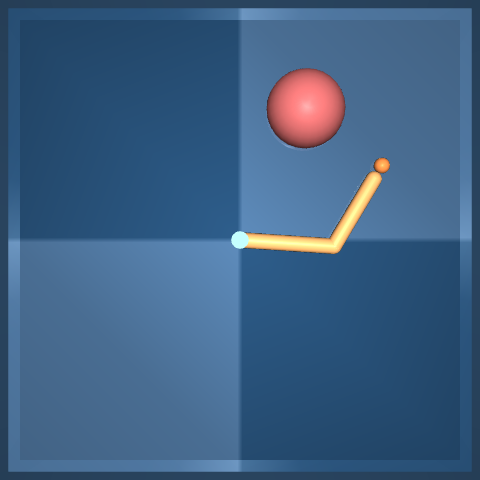}
\caption{Reacher}
\end{subfigure}\hfill%
\begin{subfigure}[t]{\width}
\includegraphics[width=\textwidth]{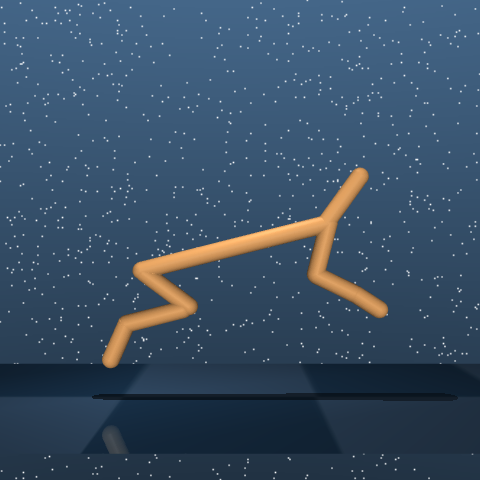}
\caption{Cheetah}
\end{subfigure}\hfill%
\begin{subfigure}[t]{\width}
\includegraphics[width=\textwidth]{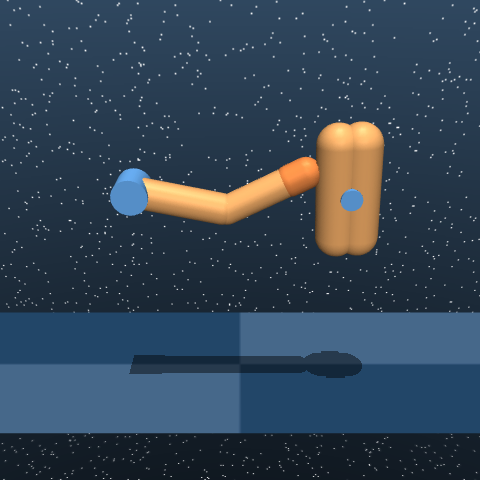}
\caption{Finger}
\end{subfigure}\hfill%
\begin{subfigure}[t]{\width}
\includegraphics[width=\textwidth]{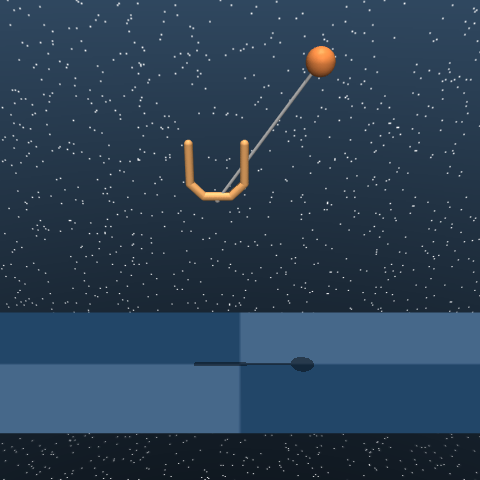}
\caption{Cup}
\end{subfigure}\hfill%
\begin{subfigure}[t]{\width}
\includegraphics[width=\textwidth]{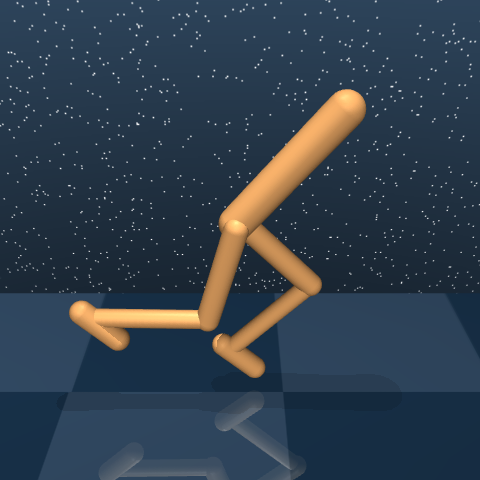}
\caption{Walker}%
\end{subfigure}%
\caption{Image-based control domains used in our experiments. The images show agent observations before downscaling to $64\times 64\times 3$ pixels.
(a) The cartpole swingup task has a fixed camera so the cart can move out of sight.
(b) The reacher task has only a sparse reward.
(c) The cheetah running task includes both contacts and a larger number of joints.
(d) The finger spinning task includes contacts between the finger and the object.
(e) The cup task has a sparse reward that is only given once the ball is caught.
(f) The walker task requires balance and predicting difficult interactions with the ground when the robot is lying down.}
\label{fig:domains}
\end{figure*}
}

Planning is a natural and powerful approach to decision making problems with known dynamics, such as game playing and simulated robot control \citep{tassa2012mpc,silver2017alphago,moravvcik2017deepstack}. To plan in unknown environments, the agent needs to learn the dynamics from experience. Learning dynamics models that are accurate enough for planning has been a long-standing challenge. 
Key difficulties include model inaccuracies, accumulating errors of multi-step predictions, failure to capture multiple possible futures, and overconfident predictions outside of the training distribution.

Planning using learned models offers several benefits over model-free reinforcement learning. First, model-based planning can be more data efficient because it leverages a richer training signal and does not require propagating rewards through Bellman backups. Moreover, planning carries the promise of increasing performance just by increasing the computational budget for searching for actions, as shown by \citet{silver2017alphago}. Finally, learned dynamics can be independent of any specific task and thus have the potential to transfer well to other tasks in the environment.

Recent work has shown promise in learning the dynamics of simple low-dimensional environments \citep{deisenroth2011pilco,gal2016deeppilco,amos2018awareness,chua2018pets,henaff2018planbybackprop}. However, these approaches typically assume access to the underlying state of the world and the reward function, which may not be available in practice. In high-dimensional environments, we would like to learn the dynamics in a compact latent space to enable fast planning. The success of such latent models has previously been limited to simple tasks such as balancing cartpoles and controlling 2-link arms from dense rewards \citep{watter2015e2c,banijamali2017rce}.

In this paper, we propose the \fullmethod (\method), a model-based agent that learns the environment dynamics from pixels and chooses actions through online planning in a compact latent space. To learn the dynamics, we use a transition model with both stochastic and deterministic components. Moreover, we experiment with a novel generalized variational objective that encourages multi-step predictions. \method solves continuous control tasks from pixels that are more difficult than those previously solved by planning with learned models.

Key contributions of this work are summarized as follows:
\begin{itemize}
\paritem{Planning in latent spaces}
We solve a variety of tasks from the DeepMind control suite, shown in \cref{fig:domains}, by learning a dynamics model and efficiently planning in its latent space. Our agent substantially outperforms the model-free A3C and in some cases D4PG algorithm in final performance, with on average $200\times$ less environment interaction and similar computation time.
\paritem{Recurrent state space model} We design a latent dynamics model with both deterministic and stochastic components \citep{buesing2018dssm,chung2015vrnn}. Our experiments indicate having both components to be crucial for high planning performance.
\paritem{Latent overshooting}
We generalize the standard variational bound to include multi-step predictions. Using only terms in latent space results in a fast regularizer that can improve long-term predictions and is compatible with any latent sequence model.
\end{itemize}

\section{Latent Space Planning}
\label{sec:planning}

\begin{algorithm}[htpb]
\SetEndCharOfAlgoLine{}
\SetKwComment{Comment}{// }{}
\SetKwInOut{Input}{Input}
\Input{\\\hspace{-3.6em}\small
\begin{tabular}[t]{l @{\hspace{.5em}} l}%
$R$ & Action repeat \\
$S$ & Seed episodes \\
$C$ & Collect interval \\
$B$ & Batch size \\
$L$ & Chunk length \\
$\alpha$ & Learning rate \\
\end{tabular}\hspace{-0.5em}%
\begin{tabular}[t]{l @{\hspace{.5em}} l}%
$\p(s_t|s_{t-1},a_{t-1})$ & Transition model \\
$\p(o_t|s_t)$ & Observation model \\
$\p(r_t|s_t)$ & Reward model \\
$\q(s_t|o_{\leq t},a_{<t})$ & Encoder \\
$\p(\epsilon)$ & Exploration noise \\
\end{tabular}%
}
\BlankLine
Initialize dataset $\mathcal{D}$ with $S$ random seed episodes. \;
Initialize model parameters $\theta$ randomly. \;
\While{not converged}{
  \BlankLine
  \Comment{Model fitting}
  \For{update step $s=1..C$}{
    Draw sequence chunks $\{(o_t,a_t,r_t)_{t=k}^{L+k}\}_{i=1}^B\sim\mathcal{D}$ uniformly at random from the dataset. \;
    Compute loss $\mathcal{L}(\theta)$ from \cref{eq:elbo}. \;
    Update model parameters $\theta\leftarrow\theta-\alpha\nabla_\theta\mathcal{L}(\theta)$. \;
  }
  \BlankLine
  \Comment{Data collection}
  $o_1\leftarrow\texttt{env.reset()}$ \;
  \For{time step $t=1..\ceil[\big]{\frac{T}{R}}$}{
    Infer belief over current state $\q(s_t|o_{\leq t},a_{<t})$ from the history. \;
    $a_t\leftarrow\texttt{planner(}\q(s_t|o_{\leq t},a_{<t}),p\texttt{)}$, see \cref{alg:planner} in the appendix for details. \;
    Add exploration noise $\epsilon\sim\p(\epsilon)$ to the action. \;
    \For{action repeat $k=1..R$}{
      $r_t^k,o_{t+1}^k\leftarrow\texttt{env.step(}a_t\texttt{)}$ \;
    }
    $r_t,o_{t+1} \leftarrow \sum_{k=1}^R r_t^k, o_{t+1}^R$ \;
  }
  $\mathcal{D}\leftarrow\mathcal{D}\cup\{(o_t,a_t,r_t)_{t=1}^T\}$ \;
  \BlankLine
}
\caption{\fullmethod (\method)}
\label{alg:agent}
\end{algorithm}

To solve unknown environments via planning, we need to model the environment dynamics from experience. \method does so by iteratively collecting data using planning and training the dynamics model on the gathered data. In this section, we introduce notation for the environment and describe the general implementation of our model-based agent. In this section, we assume access to a learned dynamics model. Our design and training objective for this model are detailed in \cref{sec:model}.

\paragraph{Problem setup}

Since individual image observations generally do not reveal the full state of the environment, we consider a partially observable Markov decision process (POMDP). We define a discrete time step~$t$, hidden states $s_t$, image observations $o_t$, continuous action vectors $a_t$, and scalar rewards $r_t$, that follow the stochastic dynamics
\begin{gather}
\begin{aligned}
\makebox[12em][l]{Transition function:} && s_t &\sim\pr[\mathrm{p}](s_t|s_{t-1},a_{t-1}) \\
\makebox[12em][l]{Observation function:} && o_t &\sim\pr[\mathrm{p}](o_t|s_t) \\
\makebox[12em][l]{Reward function:} && r_t &\sim\pr[\mathrm{p}](r_t|s_t) \\
\makebox[12em][l]{Policy:} && a_t &\sim\pr[\mathrm{p}](a_t|o_{\leq t},a_{<t}),
\label{eq:pomdp}
\end{aligned}
\raisetag{2.4\baselineskip}
\end{gather}

where we assume a fixed initial state $s_0$ without loss of generality. The goal is to implement a policy $\pr[\mathrm{p}](a_t|o_{\leq t},a_{<t})$ that maximizes the expected sum of rewards  $\E[\big]{\mathrm{p}}{\sum_{t=1}^T r_t}$, where the expectation is over the distributions of the environment and the policy.

\paragraph{Model-based planning}

\method learns a transition model $\p(s_t|s_{t-1},a_{t-1})$, observation model $\p(o_t|s_t)$, and reward model $\p(r_t|s_t)$ from previously experienced episodes (note italic letters for the model compared to upright letters for the true dynamics). The observation model provides a rich training signal but is not used for planning. We also learn an encoder $\q(s_t|o_{\leq t},a_{<t})$ to infer an approximate belief over the current hidden state from the history using filtering. Given these components, we implement the policy as a planning algorithm that searches for the best sequence of future actions. We use model-predictive control \citep[MPC;][]{richards2005mpc} to allow the agent to adapt its plan based on new observations, meaning we replan at each step. In contrast to model-free and hybrid reinforcement learning algorithms, we do not use a policy or value network.

\paragraph{Experience collection}

Since the agent may not initially visit all parts of the environment, we need to iteratively collect new experience and refine the dynamics model. We do so by planning with the partially trained model, as shown in \cref{alg:agent}. Starting from a small amount of $S$ seed episodes collected under random actions, we train the model and add one additional episode to the data set every $C$ update steps. When collecting episodes for the data set, we add small Gaussian exploration noise to the action. To reduce the planning horizon and provide a clearer learning signal to the model, we repeat each action $R$ times, as common in reinforcement learning \citep{mnih2015dqn,mnih2016a3c}.

\paragraph{Planning algorithm}

We use the cross entropy method \citep[CEM;][]{rubinstein1997cem,chua2018pets} to search for the best action sequence under the model, as outlined in \cref{alg:planner}. We decided on this algorithm because of its robustness and because it solved all considered tasks when given the true dynamics for planning. CEM is a population-based optimization algorithm that infers a distribution over action sequences that maximize the objective. As detailed in \cref{alg:planner} in the appendix, we initialize a time-dependent diagonal Gaussian belief over optimal action sequences $a_{t:t+H}\sim\Normal(\mu_{t:t+H},\sigma^2_{t:t+H}\mathbb{I})$, where $t$ is the current time step of the agent and $H$ is the length of the planning horizon. Starting from zero mean and unit variance, we repeatedly sample $J$ candidate action sequences, evaluate them under the model, and re-fit the belief to the top $K$ action sequences. After $I$ iterations, the planner returns the mean of the belief for the current time step, $\mu_t$. Importantly, after receiving the next observation, the belief over action sequences starts from zero mean and unit variance again to avoid local optima.

To evaluate a candidate action sequence under the learned model, we sample a state trajectory starting from the current state belief, and sum the mean rewards predicted along the sequence. Since we use a population-based optimizer, we found it sufficient to consider a single trajectory per action sequence and thus focus the computational budget on evaluating a larger number of different sequences. Because the reward is modeled as a function of the latent state, the planner can operate purely in latent space without generating images, which allows for fast evaluation of large batches of action sequences. The next section introduces the latent dynamics model that the planner uses.

\section{Recurrent State Space Model}
\label{sec:model}

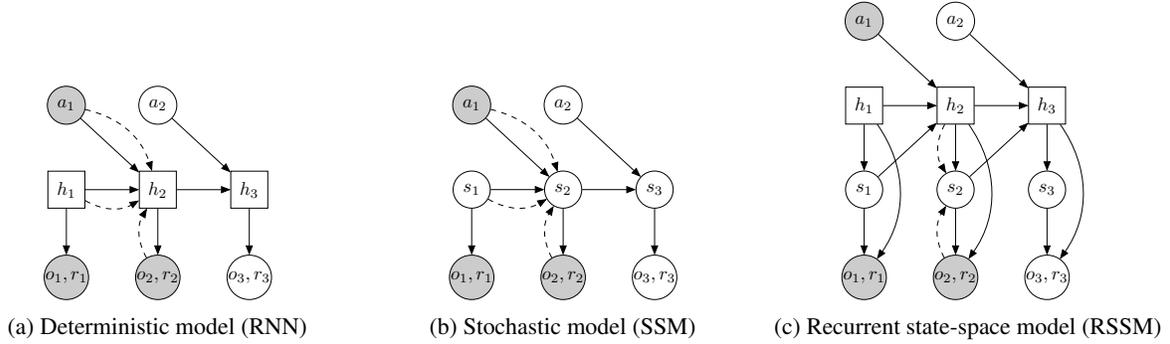
\begin{figure*}[t]
\centering
\hfil\hfil%
\begin{subfigure}[t]{.3\textwidth}
\centering
\scalebox{0.7}{%
\begin{tikzpicture}[
  node distance=2.5em, auto,
  lat/.style={draw=black, circle, minimum size=2em},
  det/.style={draw=black, rectangle, minimum size=2em},
  obs/.style={circle, draw=black, fill=black!20, minimum size=2em},
  gen/.style={->, -{Stealth[length=.5em, inset=0pt]}},
  inf/.style={dashed, ->, -{Stealth[length=.5em, inset=0pt]}},
]
\node[obs, inner sep=.02em] (o1) {$o_1,r_1$};
\node[obs, right=of o1, inner sep=.02em] (o2) {$o_2,r_2$};
\node[lat, right=of o2, inner sep=.02em] (o3) {$o_3,r_3$};
\node[det, above=of o1] (s1) {$h_1$};
\node[det, above=of o2] (s2) {$h_2$};
\node[det, above=of o3] (s3) {$h_3$};
\node[obs, above=of s1] (a1) {$a_1$};
\node[lat, above=of s2] (a2) {$a_2$};
\path (s1) edge[gen] node {} (o1);
\path (s2) edge[gen] node {} (o2);
\path (s3) edge[gen] node {} (o3);
\path (s1) edge[gen] node {} (s2);
\path (s2) edge[gen] node {} (s3);
\path (a1) edge[gen] node {} (s2);
\path (a2) edge[gen] node {} (s3);
\path (s1) edge[inf, bend right=30] node {} (s2);
\path (a1) edge[inf, bend left=30] node {} (s2);
\path (o2) edge[inf, bend left=30] node {} (s2);
\end{tikzpicture}}
\caption{Deterministic model (RNN)}
\label{fig:rnn}
\end{subfigure}\hfil%
\begin{subfigure}[t]{.3\textwidth}
\centering
\scalebox{0.7}{%
\begin{tikzpicture}[
  node distance=2.5em, auto,
  lat/.style={draw=black, circle, minimum size=2em},
  det/.style={draw=black, rectangle, minimum size=2em},
  obs/.style={circle, draw=black, fill=black!20, minimum size=2em},
  gen/.style={->, -{Stealth[length=.5em, inset=0pt]}},
  inf/.style={dashed, ->, -{Stealth[length=.5em, inset=0pt]}},
]
\node[obs, inner sep=.02em] (o1) {$o_1,r_1$};
\node[obs, right=of o1, inner sep=.02em] (o2) {$o_2,r_2$};
\node[lat, right=of o2, inner sep=.02em] (o3) {$o_3,r_3$};
\node[lat, above=of o1] (s1) {$s_1$};
\node[lat, above=of o2] (s2) {$s_2$};
\node[lat, above=of o3] (s3) {$s_3$};
\node[obs, above=of s1] (a1) {$a_1$};
\node[lat, above=of s2] (a2) {$a_2$};
\path (s1) edge[gen] node {} (o1);
\path (s2) edge[gen] node {} (o2);
\path (s3) edge[gen] node {} (o3);
\path (s1) edge[gen] node {} (s2);
\path (s2) edge[gen] node {} (s3);
\path (a1) edge[gen] node {} (s2);
\path (a2) edge[gen] node {} (s3);
\path (s1) edge[inf, bend right=30] node {} (s2);
\path (a1) edge[inf, bend left=30] node {} (s2);
\path (o2) edge[inf, bend left=30] node {} (s2);
\end{tikzpicture}}
\caption{Stochastic model (SSM)}
\label{fig:ssm}
\end{subfigure}\hfil%
\begin{subfigure}[t]{.3\textwidth}
\centering
\scalebox{0.7}{%
\begin{tikzpicture}[
  node distance=2.5em, auto,
  lat/.style={draw=black, circle, minimum size=2em},
  det/.style={draw=black, rectangle, minimum size=2em},
  obs/.style={circle, draw=black, fill=black!20, minimum size=2em},
  gen/.style={->, -{Stealth[length=.5em, inset=0pt]}},
  inf/.style={dashed, ->, -{Stealth[length=.5em, inset=0pt]}},
]
\node[obs, inner sep=.02em] (o1) {$o_1,r_1$};
\node[obs, right=of o1, inner sep=.02em] (o2) {$o_2,r_2$};
\node[lat, right=of o2, inner sep=.02em] (o3) {$o_3,r_3$};
\node[lat, above=of o1] (s1) {$s_1$};
\node[lat, above=of o2] (s2) {$s_2$};
\node[lat, above=of o3] (s3) {$s_3$};
\node[det, above=of s1] (b1) {$h_1$};
\node[det, above=of s2] (b2) {$h_2$};
\node[det, above=of s3] (b3) {$h_3$};
\node[obs, above=of b1] (a1) {$a_1$};
\node[lat, above=of b2] (a2) {$a_2$};
\path (s1) edge[gen] node {} (o1);
\path (s2) edge[gen] node {} (o2);
\path (s3) edge[gen] node {} (o3);
\path (b1) edge[gen, bend left=35] node {} (o1);
\path (b2) edge[gen, bend left=35] node {} (o2);
\path (b3) edge[gen, bend left=35] node {} (o3);
\path (b1) edge[gen] node {} (s1);
\path (b2) edge[gen] node {} (s2);
\path (b3) edge[gen] node {} (s3);
\path (b1) edge[gen] node {} (b2);
\path (b2) edge[gen] node {} (b3);
\path (s1) edge[gen] node {} (b2);
\path (s2) edge[gen] node {} (b3);
\path (a1) edge[gen] node {} (b2);
\path (a2) edge[gen] node {} (b3);
\path (b2) edge[inf, bend right=30] node {} (s2);
\path (o2) edge[inf, bend left=30] node {} (s2);
\end{tikzpicture}}
\caption{Recurrent state-space model (RSSM)}
\label{fig:rssm}
\end{subfigure}%
\hfil\hfil%
\caption{Latent dynamics model designs.
In this example, the model observes the first two time steps and predicts the third.
Circles represent stochastic variables and squares deterministic variables. Solid lines denote the generative process and dashed lines the inference model.
(a)~Transitions in a recurrent neural network are purely deterministic. This prevents the model from capturing multiple futures and makes it easy for the planner to exploit inaccuracies.
(b)~Transitions in a state-space model are purely stochastic. This makes it difficult to remember information over multiple time steps.
(c)~We split the state into stochastic and deterministic parts, allowing the model to robustly learn to predict multiple futures.}
\label{fig:models}
\end{figure*}

For planning, we need to evaluate thousands of action sequences at every time step of the agent. Therefore, we use a recurrent state-space model (RSSM) that can predict forward purely in latent space, similar to recently proposed models \citep{karl2016dvbf,buesing2018dssm,doerr2018prssm}. This model can be thought of as a non-linear Kalman filter or sequential VAE. Instead of an extensive comparison to prior architectures, we highlight two findings that can guide future designs of dynamics models: our experiments show that both stochastic and deterministic paths in the transition model are crucial for successful planning. In this section, we remind the reader of latent state-space models and then describe our dynamics model.

\paragraph{Latent dynamics}

We consider sequences $\{o_t,a_t,r_t\}_{t=1}^T$ with discrete time step $t$, image observations $o_t$, continuous action vectors $a_t$, and scalar rewards $r_t$. A typical latent state-space model is shown in \cref{fig:ssm} and resembles the structure of a partially observable Markov decision process. It defines the generative process of the images and rewards using a hidden state sequence $\{s_t\}_{t=1}^T$,
\begin{gather}
\begin{aligned}
\makebox[12em][l]{Transition model:} && s_t &\sim\p(s_t|s_{t-1},a_{t-1}) \\
\makebox[12em][l]{Observation model:} && o_t &\sim\p(o_t|s_t) \\
\makebox[12em][l]{Reward model:} && r_t &\sim\p(r_t|s_t),
\label{eq:ssm}
\end{aligned}
\raisetag{1.4\baselineskip}
\end{gather}%
where we assume a fixed initial state $s_0$ without loss of generality. The transition model is Gaussian with mean and variance parameterized by a feed-forward neural network, the observation model is Gaussian with mean parameterized by a deconvolutional neural network and identity covariance, and the reward model is a scalar Gaussian with mean parameterized by a feed-forward neural network and unit variance. Note that the log-likelihood under a Gaussian distribution with unit variance equals the mean squared error up to a constant.

\paragraph{Variational encoder}

Since the model is non-linear, we cannot directly compute the state posteriors that are needed for parameter learning. Instead, we use an encoder $\q(s_{1:T}|o_{1:T},a_{1:T})=\prod_{t=1}^T\q(s_t|s_{t-1},a_{t-1},o_t)$ to infer approximate state posteriors from past observations and actions, where $\q(s_t|s_{t-1},a_{t-1},o_t)$ is a diagonal Gaussian with mean and variance parameterized by a convolutional neural network followed by a feed-forward neural network. We use the filtering posterior that conditions on past observations since we are ultimately interested in using the model for planning, but one may also use the full smoothing posterior during training \citep{babaeizadeh2017sv2p,gregor2018tdvae}.

\paragraph{Training objective}

Using the encoder, we construct a variational bound on the data log-likelihood. For simplicity, we write losses for predicting only the observations --- the reward losses follow by analogy. The variational bound obtained using Jensen's inequality is
{\small\begin{multline}
\ln\p(o_{1:T}|a_{1:T})
\triangleq\ln\int\prod_t\p(s_t|s_{t-1},a_{t-1})\p(o_t|s_t)\d s_{1:T} \\[-2ex]
\begin{aligned}
&\geq\sum_{t=1}^T \Big(
  \describe{\E{\q(s_t|o_{\leq t},a_{<t})}{\ln\p(o_t|s_t)}}{reconstruction} \eqbr
  &\quad-\describe{\Ebelow[\big]{\q(s_{t-1}|o_{\leq t-1},a_{<t-1})}{\KL{\q(s_t|o_{\leq t},a_{<t})}{\p(s_t|s_{t-1},a_{t-1})}}}{complexity} \Big).
\end{aligned}
\label{eq:elbo}
\end{multline}}%
For the derivation, please see \cref{eq:elbo_deriv} in the appendix. Estimating the outer expectations using a single reparameterized sample yields an efficient objective for inference and learning in non-linear latent variable models that can be optimized using gradient ascent \citep{kingma2013vae,rezende2014vae,krishnan2017ssmelbo}.

\paragraph{Deterministic path}

Despite its generality, the purely stochastic transitions make it difficult for the transition model to reliably remember information for multiple time steps. In theory, this model could learn to set the variance to zero for some state components, but the optimization procedure may not find this solution. This motivates including a deterministic sequence of activation vectors $\{h_t\}_{t=1}^T$ that allow the model to access not just the last state but all previous states deterministically \citep{chung2015vrnn,buesing2018dssm}. We use such a model, shown in \cref{fig:rssm}, that we name recurrent state-space model (RSSM),
\begin{gather}
\begin{aligned}
\makebox[11em][l]{Deterministic state model:} && h_t &=f(h_{t-1},s_{t-1},a_{t-1}) \\
\makebox[11em][l]{Stochastic state model:} && s_t &\sim\p(s_t|h_t) \\
\makebox[11em][l]{Observation model:} && o_t &\sim\p(o_t|h_t,s_t) \\
\makebox[11em][l]{Reward model:} && r_t &\sim\p(r_t|h_t,s_t),
\end{aligned}
\raisetag{2.5\baselineskip}
\label{eq:rssm}
\end{gather}%
where $f(h_{t-1},s_{t-1},a_{t-1})$ is implemented as a recurrent neural network (RNN). Intuitively, we can understand this model as splitting the state into a stochastic part $s_t$ and a deterministic part $h_t$, which depend on the stochastic and deterministic parts at the previous time step through the RNN. We use the encoder $\q(s_{1:T}|o_{1:T},a_{1:T})=\prod_{t=1}^T \q(s_t|h_t,o_t)$ to parameterize the approximate state posteriors. Importantly, all information about the observations must pass through the sampling step of the encoder to avoid a deterministic shortcut from inputs to reconstructions.

In the next section, we identify a limitation of the standard objective for latent sequence models and propose a generalization of it that improves long-term predictions.

\section{Latent Overshooting}
\label{sec:overshooting}

\begin{figure*}[t]
\centering
\hfil\hfil%
\begin{subfigure}[t]{.27\textwidth}
\centering
\scalebox{0.7}{%
\begin{tikzpicture}[
  node distance=2.5em, auto,
  lat/.style={draw=black, circle, minimum size=2em},
  obs/.style={circle, draw=black, fill=black!20, minimum size=2em},
  gen/.style={->, -{Stealth[length=.5em, inset=0pt]}},
  inf/.style={dashed, ->, -{Stealth[length=.5em, inset=0pt]}},
  div/.style={decorate, decoration={snake, amplitude=.15em, segment length=.8em, post length=.45em}, ->, -{Stealth[length=.5em, inset=0pt]}},
]
\node[obs, inner sep=.02em] (o1) {$o_1,r_1$};
\node[obs, right=of o1, inner sep=.02em] (o2) {$o_2,r_2$};
\node[obs, right=of o2, inner sep=.02em] (o3) {$o_3,r_3$};
\node[lat, above=of o1] (q11) {$s_{1|1}$};
\node[lat, above=of o2] (q22) {$s_{2|2}$};
\node[lat, above=of o3] (q33) {$s_{3|3}$};
\node[lat, above=of q22] (q21) {$s_{2|1}$};
\node[lat, above=of q33] (q32) {$s_{3|2}$};
\path (q11) edge[gen] node {} (o1);
\path (q22) edge[gen] node {} (o2);
\path (q33) edge[gen] node {} (o3);
\path (q11) edge[gen] node {} (q21);
\path (q22) edge[gen] node {} (q32);
\path (o1) edge[inf, bend left=40] node {} (q11);
\path (o2) edge[inf, bend left=40] node {} (q22);
\path (o3) edge[inf, bend left=40] node {} (q33);
\path (q11) edge[inf] node {} (q22);
\path (q22) edge[inf] node {} (q33);
\path (q22) edge[div] node {} (q21);
\path (q33) edge[div] node {} (q32);
\end{tikzpicture}}
\caption{Standard variational bound}
\label{fig:elbo}
\end{subfigure}\hfil%
\begin{subfigure}[t]{.27\textwidth}
\centering
\scalebox{0.7}{%
\begin{tikzpicture}[
  node distance=2.5em, auto,
  lat/.style={draw=black, circle, minimum size=2em},
  obs/.style={circle, draw=black, fill=black!20, minimum size=2em},
  gen/.style={->, -{Stealth[length=.5em, inset=0pt]}},
  inf/.style={dashed, ->, -{Stealth[length=.5em, inset=0pt]}},
  div/.style={decorate, decoration={snake, amplitude=.15em, segment length=.8em, post length=.45em}, ->, -{Stealth[length=.5em, inset=0pt]}},
]
\node[obs, inner sep=.02em] (o1) {$o_1,r_1$};
\node[obs, right=of o1, inner sep=.02em] (o2) {$o_2,r_2$};
\node[obs, right=of o2, inner sep=.02em] (o3) {$o_3,r_3$};
\node[lat, above=of o1] (q11) {$s_{1|1}$};
\node[lat, above=of o2] (q22) {$s_{2|2}$};
\node[lat, above=of o3] (q33) {$s_{3|3}$};
\node[lat, above=of q22] (q21) {$s_{2|1}$};
\node[lat, above=of q33] (q32) {$s_{3|2}$};
\node[lat, above=of q32] (q31) {$s_{3|1}$};
\path (q11) edge[gen] node {} (o1);
\path (q22) edge[gen] node {} (o2);
\path (q33) edge[gen] node {} (o3);
\path (q11) edge[gen] node {} (q21);
\path (q22) edge[gen] node {} (q32);
\path (q21) edge[gen] node {} (q31);
\path (q21) edge[gen, bend left=30] node {} (o2);
\path (q32) edge[gen, bend left=30] node {} (o3);
\path (q31) edge[gen, bend left=50] node {} (o3);
\path (o1) edge[inf, bend left=40] node {} (q11);
\path (o2) edge[inf, bend left=40] node {} (q22);
\path (o3) edge[inf, bend left=40] node {} (q33);
\path (q11) edge[inf] node {} (q22);
\path (q22) edge[inf] node {} (q33);
\end{tikzpicture}}
\caption{Observation overshooting}
\label{fig:obsov}
\end{subfigure}\hfil%
\begin{subfigure}[t]{.27\textwidth}
\centering
\scalebox{0.7}{%
\begin{tikzpicture}[
  node distance=2.5em, auto,
  lat/.style={draw=black, circle, minimum size=2em},
  obs/.style={circle, draw=black, fill=black!20, minimum size=2em},
  gen/.style={->, -{Stealth[length=.5em, inset=0pt]}},
  inf/.style={dashed, ->, -{Stealth[length=.5em, inset=0pt]}},
  div/.style={decorate, decoration={snake, amplitude=.15em, segment length=.8em, post length=.45em}, ->, -{Stealth[length=.5em, inset=0pt]}},
]
\node[obs, inner sep=.02em] (o1) {$o_1,r_1$};
\node[obs, right=of o1, inner sep=.02em] (o2) {$o_2,r_2$};
\node[obs, right=of o2, inner sep=.02em] (o3) {$o_3,r_3$};
\node[lat, above=of o1] (q11) {$s_{1|1}$};
\node[lat, above=of o2] (q22) {$s_{2|2}$};
\node[lat, above=of o3] (q33) {$s_{3|3}$};
\node[lat, above=of q22] (q21) {$s_{2|1}$};
\node[lat, above=of q33] (q32) {$s_{3|2}$};
\node[lat, above=of q32] (q31) {$s_{3|1}$};
\path (q11) edge[gen] node {} (o1);
\path (q22) edge[gen] node {} (o2);
\path (q33) edge[gen] node {} (o3);
\path (q11) edge[gen] node {} (q21);
\path (q22) edge[gen] node {} (q32);
\path (q21) edge[gen] node {} (q31);
\path (o1) edge[inf, bend left=40] node {} (q11);
\path (o2) edge[inf, bend left=40] node {} (q22);
\path (o3) edge[inf, bend left=40] node {} (q33);
\path (q11) edge[inf] node {} (q22);
\path (q22) edge[inf] node {} (q33);
\path (q22) edge[div] node {} (q21);
\path (q33) edge[div] node {} (q32);
\path (q33) edge[div, bend right=40] node {} (q31);
\end{tikzpicture}}
\caption{Latent overshooting}
\label{fig:latov}
\end{subfigure}%
\hfil\hfil%
\caption{Unrolling schemes.
The labels $s_{i|j}$ are short for the state at time $i$ conditioned on observations up to time $j$.
Arrows pointing at shaded circles indicate log-likelihood loss terms. Wavy arrows indicate KL-divergence loss terms.
(a) The standard variational objectives decodes the posterior at every step to compute the reconstruction loss. It also places a KL on the prior and posterior at every step, which trains the transition function for one-step predictions.
(b) Observation overshooting \citep{amos2018awareness} decodes all multi-step predictions to apply additional reconstruction losses. This is typically too expensive in image domains.
(c) Latent overshooting predicts all multi-step priors. These state beliefs are trained towards their corresponding posteriors in latent space to encourage accurate multi-step predictions.}
\label{fig:overshooting}
\end{figure*}
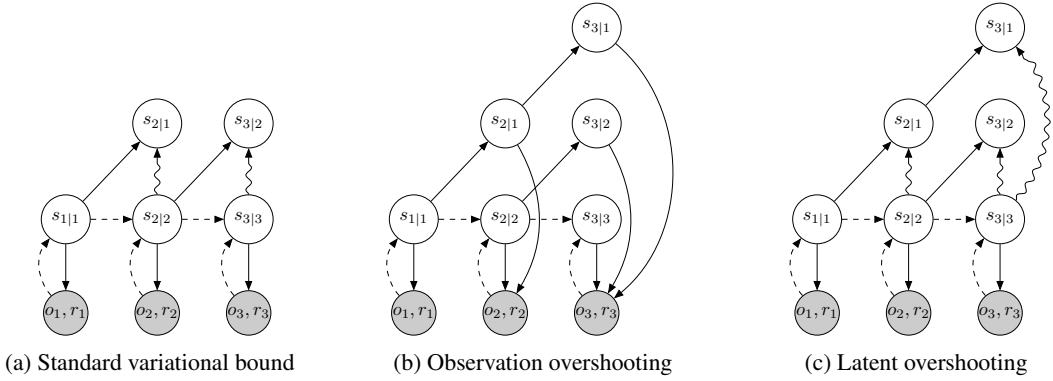

In the previous section, we derived the typical variational bound for learning and inference in latent sequence models (\cref{eq:elbo}). As show in \cref{fig:elbo}, this objective function contains reconstruction terms for the observations and KL-divergence regularizers for the approximate posteriors. A limitation of this objective is that the stochastic path of the transition function $\p(s_t|s_{t-1},a_{t-1})$ is only trained via the KL-divergence regularizers for one-step predictions: the gradient flows through $\p(s_t|s_{t-1},a_{t-1})$ directly into $\q(s_{t-1})$ but never traverses a chain of multiple $\p(s_t|s_{t-1},a_{t-1})$. In this section, we generalize this variational bound to \emph{latent overshooting}, which trains all multi-step predictions in latent space. We found that several dynamics models benefit from latent overshooting, although our final agent using the RSSM model does not require it (see \cref{sec:exp_overshooting}).

\paragraph{Limited capacity}

If we could train our model to make perfect one-step predictions, it would also make perfect multi-step predictions, so this would not be a problem. However, when using a model with limited capacity and restricted distributional family, training the model only on one-step predictions until convergence does in general not coincide with the model that is best at multi-step predictions. For successful planning, we need accurate multi-step predictions. Therefore, we take inspiration from \citet{amos2018awareness} and earlier related ideas \citep{krishnan2015deepkalman,lamb2016professor,chiappa2017recurrent}, and train the model on multi-step predictions of all distances. We develop this idea for latent sequence models, showing that multi-step predictions can be improved by a loss in latent space, without having to generate additional images.

\paragraph{Multi-step prediction}

We start by generalizing the standard variational bound (\cref{eq:elbo}) from training one-step predictions to training multi-step predictions of a fixed distance $d$. For ease of notation, we omit actions in the conditioning set here; every distribution over $s_t$ is conditioned upon $a_{<t}$. We first define multi-step predictions, which are computed by repeatedly applying the transition model and integrating out the intermediate states,
\begin{equation}
\begin{aligned}
\p(s_t|s_{t-d})
&\triangleq\int\prod_{\tau=t-d+1}^t \p(s_\tau|s_{\tau-1})\d s_{t-d+1:t-1} \\
&=\E{\p(s_{t-1}|s_{t-d})}{\p(s_t|s_{t-1})}.
\end{aligned}
\end{equation}%
The case $d=1$ recovers the one-step transitions used in the original model. Given this definition of a multi-step prediction, we generalize \cref{eq:elbo} to the variational bound on the multi-step predictive distribution $p_d$,
\begin{equation}
\begin{aligned}
&\ln\pr[p_d](o_{1:T})
\triangleq\ln\int\prod_{t=1}^T\p(s_t|s_{t-d})\p(o_t|s_t)\d s_{1:T} \\[-2ex]
&\geq\sum_{t=1}^T \Big(
  \describe{\E{\q(s_t|o_{\leq t})}{\ln\p(o_t|s_t)}}{reconstruction} \eqbr&\quad
  -\describe{\Ebelow[\big]{\p(s_{t-1}|s_{t-d})\q(s_{t-d}|o_{\leq t-d})}{\KL{\q(s_t|o_{\leq t})}{\p(s_t|s_{t-1})}}}{multi-step prediction} \Big).
\end{aligned}
\label{eq:dstep}
\end{equation}%
For the derivation, please see \cref{eq:dstep_deriv} in the appendix. Maximizing this objective trains the multi-step predictive distribution. This reflects the fact that during planning, the model makes predictions without having access to all the preceding observations.

We conjecture that \cref{eq:dstep} is also a lower bound on $\ln\p(o_{1:T})$ based on the data processing inequality. Since the latent state sequence is Markovian, for $d\geq 1$ we have $\MI{s_t;s_{t-d}}\leq\MI{s_t;s_{t-1}}$ and thus
$\E{}{\ln\pr[p_d](o_{1:T})}\leq\E{}{\ln\p(o_{1:T})}$. Hence, every bound on the multi-step predictive distribution is also a bound on the one-step predictive distribution in expectation over the data set. For details, please see \cref{eq:dataproc_deriv} in the appendix. In the next paragraph, we alleviate the limitation that a particular $p_d$ only trains predictions of one distance and arrive at our final objective.

\paragraph{Latent overshooting}

We introduced a bound on predictions of a given distance $d$. However, for planning we need accurate predictions not just for a fixed distance but for all distances up to the planning horizon. We introduce latent overshooting for this, an objective function for latent sequence models that generalizes the standard variational bound (\cref{eq:elbo}) to train the model on multi-step predictions of all distances $1 \leq d \leq D$,
{\small\begin{multline}
\frac{1}{D}\sum_{d=1}^D\ln\pr[p_d](o_{1:T}) \geq
\sum_{t=1}^T \Big(
  \describe{\E{\q(s_t|o_{\leq t})}{\ln\p(o_t|s_t)}}{reconstruction} \eqbr
  -\describe{\frac{1}{D}\sum_{d=1}^D
    \beta_d\Ebelow[\big]{\p(s_{t-1}|s_{t-d})\q(s_{t-d}|o_{\leq t-d})}{\KL{\q(s_t|o_{\leq t})}{\p(s_t|s_{t-1})}}}{latent overshooting} \Big).
\label{eq:latov}
\end{multline}}%
Latent overshooting can be interpreted as a regularizer in latent space that encourages consistency between one-step and multi-step predictions, which we know should be equivalent in expectation over the data set. We include weighting factors $\{\beta_d\}_{d=1}^D$ analogously to the $\beta$-VAE \citep{higgins2016beta}. While we set all $\beta_{>1}$ to the same value for simplicity, they could be chosen to let the model focus more on long-term or short-term predictions. In practice, we stop gradients of the posterior distributions for overshooting distances $d>1$, so that the multi-step predictions are trained towards the informed posteriors, but not the other way around.

\section{Experiments}
\label{sec:control}

\begin{figure*}[t]
\providecommand{\width}{}
\renewcommand{\width}{0.28\textwidth}
\providecommand{\path}{}
\renewcommand{\path}{score_model}
\captionsetup[subfigure]{position=above,labelformat=empty,margin={1em,0em},skip=-.1em}
\centering\hfil\hfil%
\begin{subfigure}[t]{\width}
\centering
\caption{Cartpole Swing Up}
\includegraphics[width=\textwidth]{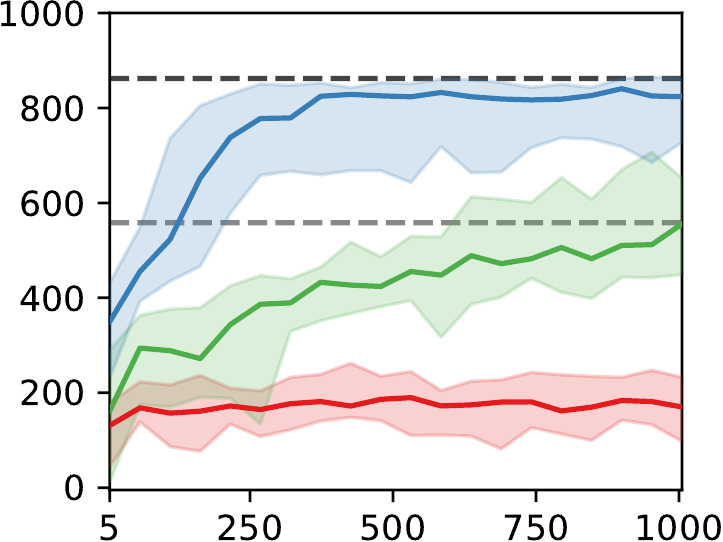}
\end{subfigure}\hfil%
\begin{subfigure}[t]{\width}
\centering
\caption{Reacher Easy}
\includegraphics[width=\textwidth]{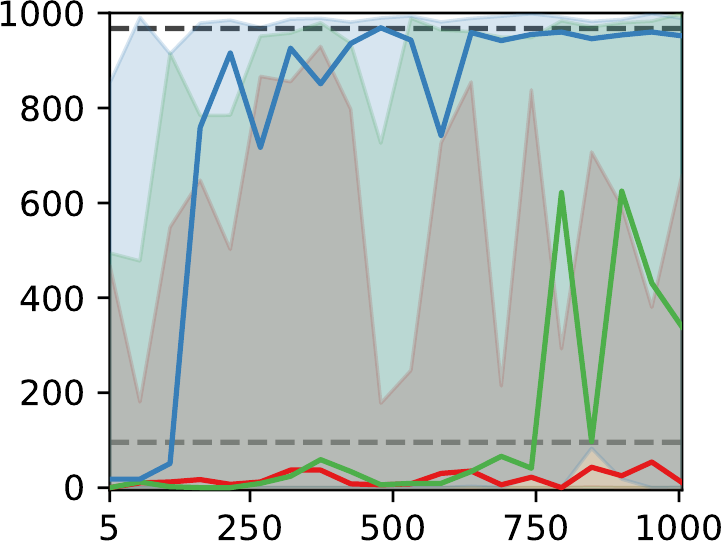}
\end{subfigure}\hfil%
\begin{subfigure}[t]{\width}
\centering
\caption{Cheetah Run}
\includegraphics[width=\textwidth]{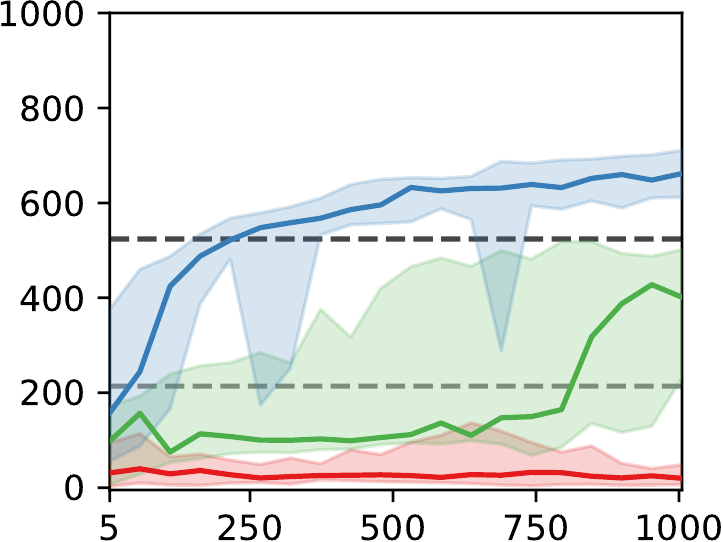}
\end{subfigure}\hfil\hfil\hfil \\
\vspace{.7em}\hfil\hfil%
\begin{subfigure}[t]{\width}
\centering
\caption{Finger Spin}
\includegraphics[width=\textwidth]{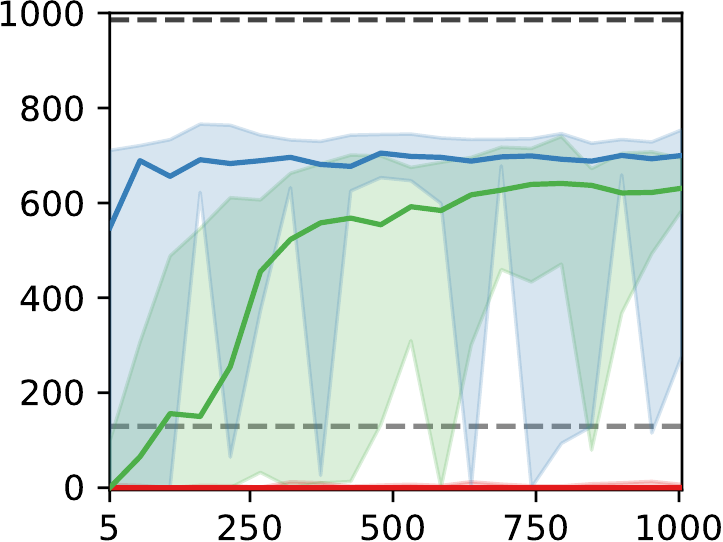}
\end{subfigure}\hfil%
\begin{subfigure}[t]{\width}
\centering
\caption{Cup Catch}
\includegraphics[width=\textwidth]{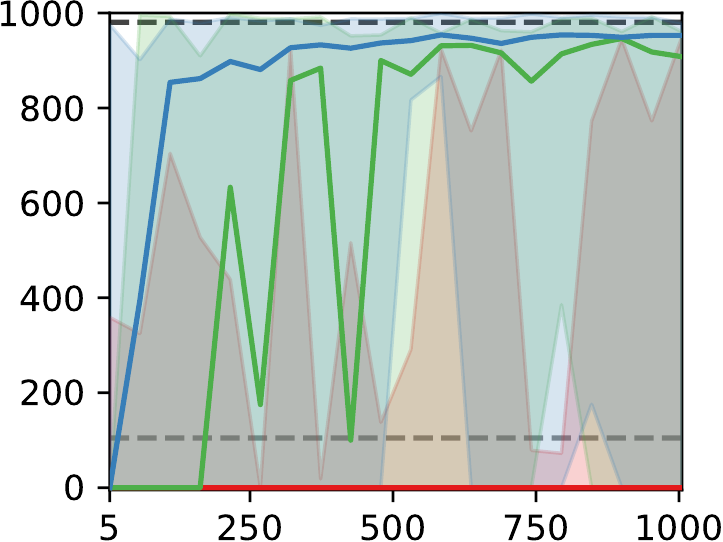}
\end{subfigure}\hfil%
\begin{subfigure}[t]{\width}
\centering
\caption{Walker Walk}
\includegraphics[width=\textwidth]{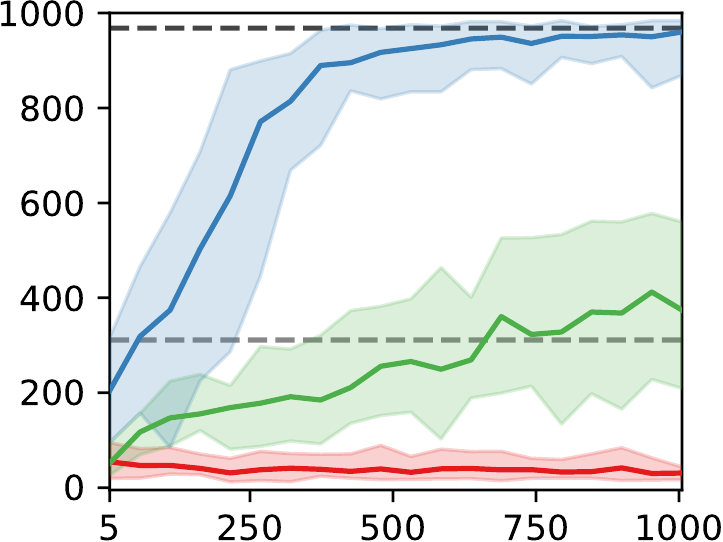}
\end{subfigure}\hfil\hfil\hfil \\
\vspace{1em}\hspace{1em}
\begin{subfigure}[t]{.75\textwidth}
\centering
\includegraphics[width=\textwidth]{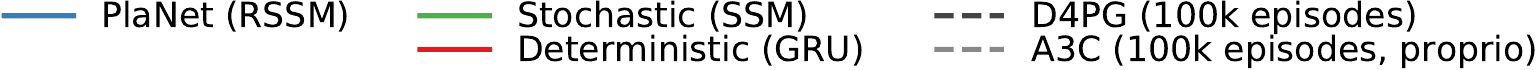}
\end{subfigure}\hfil%
\caption{Comparison of \method to model-free algorithms and other model designs. Plots show test performance over the number of collected episodes. We compare \method using our RSSM (\cref{sec:model}) to purely deterministic (GRU) and purely stochastic models (SSM). The RNN does not use latent overshooting, as it does not have stochastic latents. The lines show medians and the areas show percentiles 5 to 95 over 5 seeds and 10 trajectories. The shaded areas are large on two of the tasks due to the sparse rewards.}
\label{fig:score_model}
\end{figure*}

\begin{figure*}[t]
\providecommand{\width}{}
\renewcommand{\width}{0.28\textwidth}
\providecommand{\path}{}
\renewcommand{\path}{score_agent}
\captionsetup[subfigure]{position=above,labelformat=empty,margin={1em,0em},skip=-.1em}
\centering\hfil\hfil%
\begin{subfigure}[t]{\width}
\centering
\caption{Cartpole Swing Up}
\includegraphics[width=\textwidth]{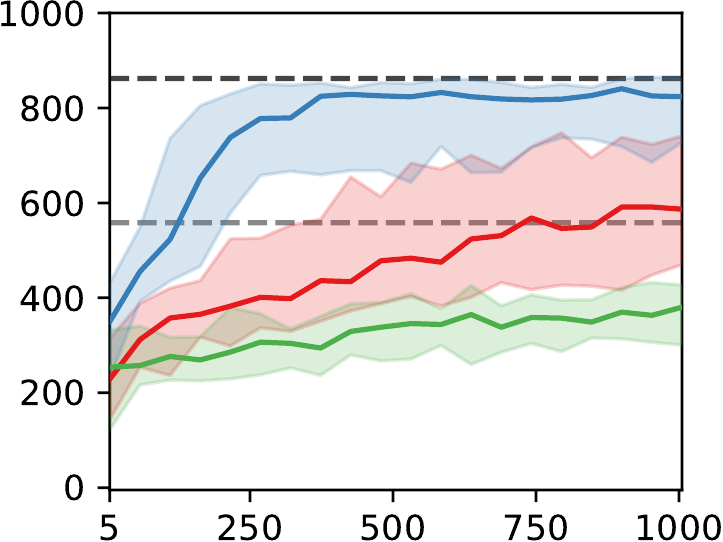}
\end{subfigure}\hfil%
\begin{subfigure}[t]{\width}
\centering
\caption{Reacher Easy}
\includegraphics[width=\textwidth]{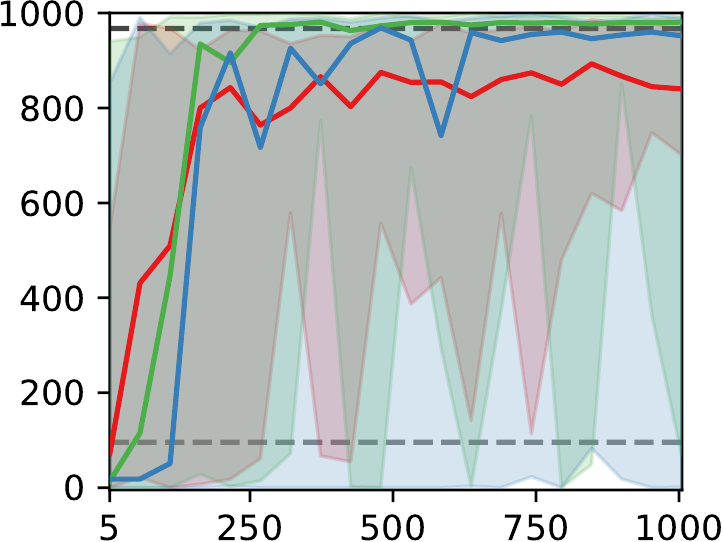}
\end{subfigure}\hfil%
\begin{subfigure}[t]{\width}
\centering
\caption{Cheetah Run}
\includegraphics[width=\textwidth]{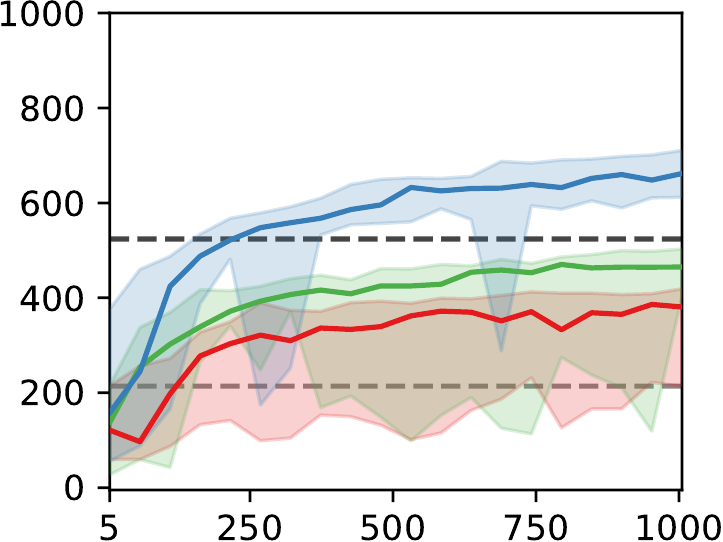}
\end{subfigure}\hfil\hfil\hfil \\
\vspace{.7em}\hfil\hfil%
\begin{subfigure}[t]{\width}
\centering
\caption{Finger Spin}
\includegraphics[width=\textwidth]{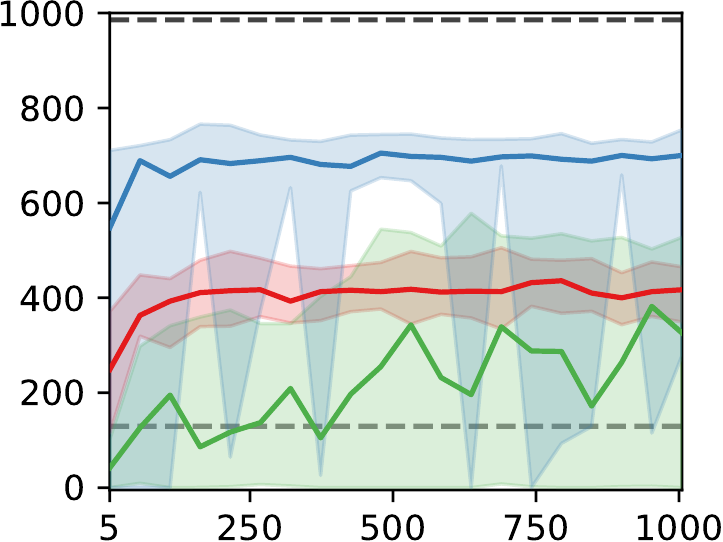}
\end{subfigure}\hfil%
\begin{subfigure}[t]{\width}
\centering
\caption{Cup Catch}
\includegraphics[width=\textwidth]{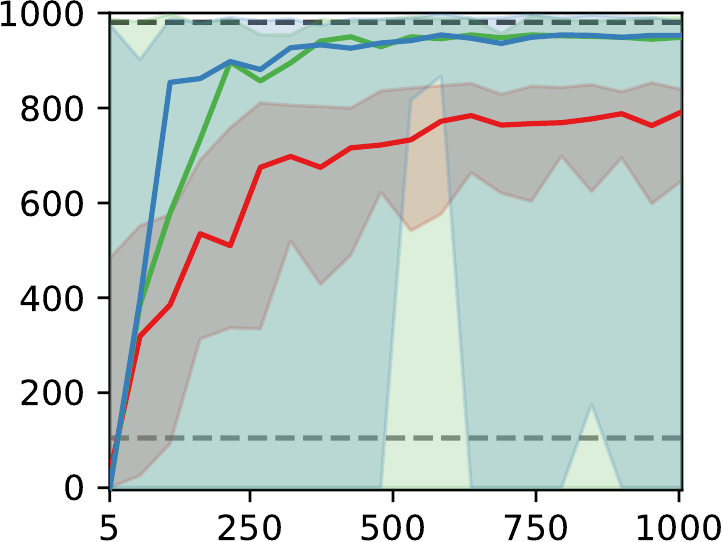}
\end{subfigure}\hfil%
\begin{subfigure}[t]{\width}
\centering
\caption{Walker Walk}
\includegraphics[width=\textwidth]{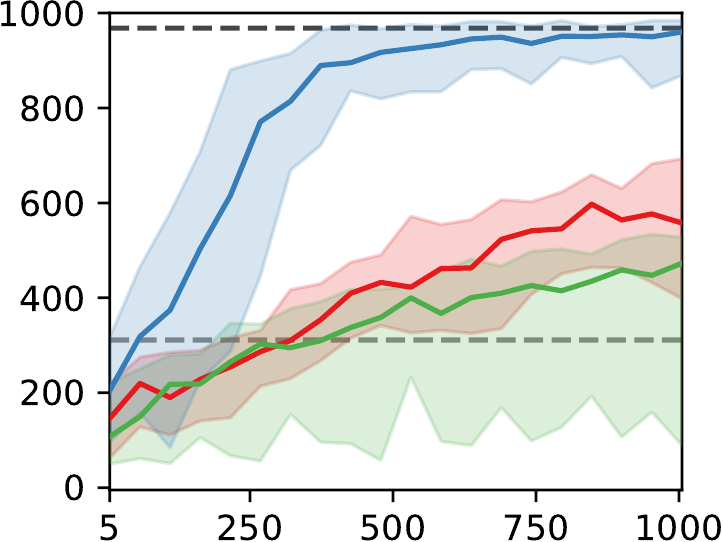}
\end{subfigure}\hfil\hfil\hfil \\
\vspace{1em}\hspace{1em}
\begin{subfigure}[t]{.7\textwidth}
\centering
\includegraphics[width=\textwidth]{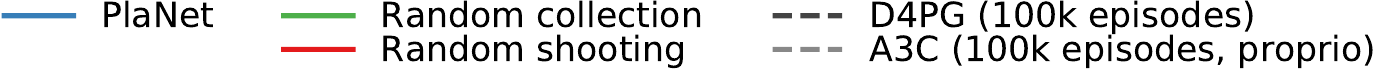}
\end{subfigure}\hfil%
\caption{Comparison of agent designs. Plots show test performance over the number of collected episodes. We compare \method, a version that collects data under random actions (random collection), and a version that chooses the best action out of $1000$ sequences at each environment step (random shooting) without iteratively refining plans via CEM. The lines show medians and the areas show percentiles 5 to 95 over 5 seeds and 10 trajectories.}
\label{fig:score_agent}
\end{figure*}

We evaluate \method on six continuous control tasks from pixels. We explore multiple design axes of the agent: the stochastic and deterministic paths in the dynamics model, iterative planning, and online experience collection. We refer to the appendix for hyper parameters (\cref{sec:hparams}) and additional experiments (\cref{sec:exp_overshooting,sec:exp_activation,sec:exp_multi}). Besides the action repeat, we use the same hyper parameters for all tasks. Within less than one hundredth the episodes, \method outperforms A3C \citep{mnih2016a3c} and achieves similar performance to the top model-free algorithm D4PG \citep{barth2018d4pg}. The training time of 10 to 20 hours (depending on the task) on a single Nvidia V100 GPU compares favorably to that of A3C and D4PG. Our implementation uses TensorFlow Probability \citep{dillon2017tfd}. Please visit \,\url{https://danijar.com/planet}\, for access to the code and videos of the trained agent.

For our evaluation, we consider six image-based continuous control tasks of the DeepMind control suite \citep{tassa2018dmcontrol}, shown in \Cref{fig:domains}. These environments provide qualitatively different challenges. The cartpole swingup task requires a long planning horizon and to memorize the cart when it is out of view, reacher has a sparse reward given when the hand and goal area overlap, finger spinning includes contact dynamics between the finger and the object, cheetah exhibits larger state and action spaces, the cup task only has a sparse reward for when the ball is caught, and the walker is challenging because the robot first has to stand up and then walk, resulting in collisions with the ground that are difficult to predict. In all tasks, the only observations are third-person camera images of size $64\times 64\times 3$ pixels.

\paragraph{Comparison to model-free methods}

\cref{fig:score_model} compares the performance of \method to the model-free algorithms reported by \citet{tassa2018dmcontrol}. Within 100 episodes, \method outperforms the policy-gradient method A3C trained from proprioceptive states for 100,000 episodes, on all tasks. After 500 episodes, it achieves performance similar to D4PG, trained from images for 100,000 episodes, except for the finger task. \method surpasses the final performance of D4PG with a relative improvement of $26\%$ on the cheetah running task. We refer to \cref{tab:control} for numerical results, which also includes the performance of CEM planning with the true dynamics of the simulator.

\paragraph{Model designs}

\cref{fig:score_model} additionally compares design choices of the dynamics model. We train \method using our recurrent state-space model (RSSM), as well as versions with purely deterministic GRU \citep{cho2014gru}, and purely stochastic state-space model (SSM). We observe the importance of both stochastic and deterministic elements in the transition function on all tasks. The deterministic part allows the model to remember information over many time steps. The stochastic component is even more important -- the agent does not learn without it. This could be because the tasks are stochastic from the agent's perspective due to partial observability of the initial states. The noise might also add a safety margin to the planning objective that results in more robust action sequences.

\paragraph{Agent designs}

\cref{fig:score_agent} compares \method, a version collecting episodes under random actions rather than by planning, and a version that at each environment step selects the best action out of $1000$ sequences rather than iteratively refining plans via CEM. We observe that online data collection helps for all tasks and is necessary for the cartpole, finger, and walker tasks. Iterative search for action sequences using CEM improves performance on all tasks.

\paragraph{One agent all tasks}

\cref{fig:score_multi} in the appendix shows the performance of a single agent trained on all six tasks. The agent is not told which task it is facing; it needs to infer this from the image observations. We pad the action spaces with unused elements to make them compatible and adapt \cref{alg:agent} to collect one episode of each task every $C$ update steps. We use the same hyper parameters as for the main experiments above. The agent solves all tasks while learning slower compared to individually trained agents. This indicates that the model can learn to predict multiple domains, regardless of the conceptually different visuals.

\begin{table*}[!tb]
\newcommand{\factor}{}
\small
\centering
\caption{Comparison of \method to the model-free algorithms A3C and D4PG reported by \citet{tassa2018dmcontrol}. The training curves for these are shown as orange lines in Figure~4 and as solid green lines in Figure~6 in their paper. From these, we estimate the number of episodes that D4PG takes to achieve the final performance of \method to estimate the data efficiency gain. We further include CEM planning ($H=12,I=10,J=1000,K=100$) with the true simulator instead of learned dynamics as an estimated upper bound on performance. Numbers indicate mean final performance over 5 seeds and 10 trajectories.}
\label{tab:control}
\begin{tabular}{llrrrrrrr}
\toprule
\textbf{Method} & \textbf{Modality} & \textbf{Episodes} & \textbf{\rot{Cartpole\\Swing Up}} & \textbf{\rot{Reacher\\Easy}} & \textbf{\rot{Cheetah\\Run}} & \textbf{\rot{Finger\\Spin}} & \textbf{\rot{Cup\\Catch}} & \textbf{\rot{Walker\\Walk}} \\ \midrule
A3C                  & proprioceptive  & 100,000 & 558 & 285 & 214 & 129 & 105 & 311 \\
D4PG                 & pixels          & 100,000 & 862 & 967 & 524 & 985 & 980 & 968 \\
\method (ours)       & pixels          & 1,000   & 821 & 832 & 662 & 700 & 930 & 951 \\
CEM + true simulator & simulator state & 0       & 850 & 964 & 656 & 825 & 993 & 994 \\
\midrule
\multicolumn{3}{l}{Data efficiency gain PlaNet over D4PG (factor)} & 250\factor & 40\factor & 500+\factor & 300\factor & 100\factor & 90\factor \\
\bottomrule
\end{tabular}
\end{table*}

\section{Related Work}
\label{related_work}

Previous work in model-based reinforcement learning has focused on planning in
low-dimensional state spaces \linebreak \citep{gal2016deeppilco,higuera2018synthesizing,henaff2018planbybackprop,chua2018pets}, combining the benefits of model-based and model-free approaches \citep{kalweit2017blending,nagabandi2017mbmf,weber2017i2a,kurutach2018modeltrpo,buckman2018steve,ha2018worldmodels,wayne2018merlin,igl2018dvrl,srinivas2018upn}, and pure video prediction without planning \citep{oh2015atari,krishnan2015deepkalman,karl2016dvbf,chiappa2017recurrent,babaeizadeh2017sv2p,gemici2017temporalmemory,denton2018stochastic,buesing2018dssm,doerr2018prssm,gregor2018tdvae}. \Cref{sec:more_related_work} reviews these orthogonal research directions in more detail.

Relatively few works have demonstrated successful planning from pixels using learned dynamics models. The robotics community focuses on video prediction models for planning \citep{agrawal2016poking,finn2017foresight,ebert2018foresight,zhang2018solar} that deal with the visual complexity of the real world and solve tasks with a simple gripper, such as grasping or pushing objects. In comparison, we focus on simulated environments, where we leverage latent planning to scale to larger state and action spaces, longer planning horizons, as well as sparse reward tasks. E2C \citep{watter2015e2c} and RCE \citep{banijamali2017rce} embed images into a latent space, where they learn local-linear latent transitions and plan for actions using LQR. These methods balance simulated cartpoles and control 2-link arms from images, but have been difficult to scale up. We lift the Markov assumption of these models, making our method applicable under partial observability, and present results on more challenging environments that include longer planning horizons, contact dynamics, and sparse rewards.

\section{Discussion}
\label{discussion}

We present \method, a model-based agent that learns a latent dynamics model from image observations and chooses actions by fast planning in latent space. To enable accurate long-term predictions, we design a model with both stochastic and deterministic paths. We show that our agent succeeds at several continuous control tasks from image observations, reaching performance that is comparable to the best model-free algorithms while using $200\times$ fewer episodes and similar or less computation time. The results show that learning latent dynamics models for planning in image domains is a promising approach.

Directions for future work include learning temporal abstraction instead of using a fixed action repeat, possibly through hierarchical models. To further improve final performance, one could learn a value function to approximate the sum of rewards beyond the planning horizon. Moreover, gradient-based planning could increase the computational efficiency of the agent and learning representations without reconstruction could help to solve tasks with higher visual diversity. Our work provides a starting point for multi-task control by sharing the dynamics model.
\paragraph{Acknowledgements}

\begin{hyphenrules}{nohyphenation}
We thank Jacob Buckman, Nicolas Heess, John Schulman, Rishabh Agarwal, Silviu Pitis, Mohammad Norouzi, George Tucker, David Duvenaud, Shane Gu, Chelsea Finn, Steven Bohez, Jimmy Ba, Stephanie Chan, and Jenny Liu for helpful discussions.
\end{hyphenrules}

\clearpage
\bibliography{references}

\clearpage
\onecolumn
\appendix
\section{Hyper Parameters}
\label{sec:hparams}

We use the convolutional and deconvolutional networks from \citet{ha2018worldmodels}, a GRU \citep{cho2014gru} with $200$ units as deterministic path in the dynamics model, and implement all other functions as two fully connected layers of size $200$ with ReLU activations \citep{nair2010relu}. Distributions in latent space are $30$-dimensional diagonal Gaussians with predicted mean and standard deviation.

We pre-process images by reducing the bit depth to 5 bits as in \citet{kingma2018glow}. The model is trained using the Adam optimizer \citep{kingma2014adam} with a learning rate of $10^{-3}$, $\epsilon=10^{-4}$, and gradient clipping norm of $1000$ on batches of $B=50$ sequence chunks of length $L=50$. We do not scale the KL divergence terms relatively to the reconstruction terms but grant the model $3$ free nats by clipping the divergence loss below this value. In a previous version of the agent, we used latent overshooting and an additional fixed global prior, but we found this to not be necessary.

For planning, we use CEM with horizon length $H=12$, optimization iterations $I=10$, candidate samples $J=1000$, and refitting to the best $K=100$. We start from $S=5$ seed episodes with random actions and collect another episode every $C=100$ update steps under $\epsilon\sim\Normal(0,0.3)$ action noise. The action repeat differs between domains: cartpole ($R=8$), reacher ($R=4$), cheetah ($R=4$), finger ($R=2$), cup ($R=4$), walker ($R=2$). We found important hyper parameters to be the action repeat, the KL-divergence scales $\beta$, and the learning rate.

\section{Planning Algorithm}
\begin{algorithm}[h!]
\SetEndCharOfAlgoLine{}
\SetKwComment{Comment}{// }{}
\SetKwInOut{Input}{Input}
\Input{
\begin{tabular}[t]{l @{\hspace{.5em}} l}
$H$ & Planning horizon distance \\
$I$ & Optimization iterations \\
$J$ & Candidates per iteration \\
$K$ & Number of top candidates to fit \\
\end{tabular}%
\begin{tabular}[t]{l @{\hspace{.5em}} l}
$\q(s_t|o_{\leq t},a_{<t})$ & Current state belief \\
$\p(s_t|s_{t-1},a_{t-1})$ & Transition model \\
$\p(r_t|s_t)$ & Reward model \\
\end{tabular}%
}
\BlankLine
Initialize factorized belief over action sequences $\q(a_{t:t+H})\leftarrow\Normal(0,\I)$. \;
\For{optimization iteration $i=1..I$}{
  \Comment{Evaluate $J$ action sequences from the current belief.}
  \For{candidate action sequence $j=1..J$}{
      $a^{(j)}_{t:t+H}\sim\q(a_{t:t+H})$ \;
      $s^{(j)}_{t:t+H+1}\sim
        \q(s_t|o_{1:t},a_{1:t-1})
        \prod_{\tau=t+1}^{t+H+1}\p(s_\tau|s_{\tau-1},a^{(j)}_{\tau-1})$ \;
      $R^{(j)}=\sum_{\tau=t+1}^{t+H+1}\E{}{\p(r_\tau|s^{(j)}_\tau)}$ \;
  }
  \BlankLine
  \Comment{Re-fit belief to the $K$ best action sequences.}
  $\mathcal{K}\leftarrow\mathrm{argsort}(\{R^{(j)}\}_{j=1}^J)_{1:K}$ \;
  $\mu_{t:t+H}=\frac{1}{K}\sum_{k\in\mathcal{K}} a_{t:t+H}^{(k)}, \quad
  \sigma_{t:t+H}=\frac{1}{K-1}\sum_{k\in\mathcal{K}}|a_{t:t+H}^{(k)}-\mu_{t:t+H}|$. \;
  $\q(a_{t:t+H})\leftarrow\Normal(\mu_{t:t+H},\sigma_{t:t+H}^2\I)$ \;
}
\Return{first action mean $\mu_t$.}
\BlankLine
\caption{Latent planning with CEM}
\label{alg:planner}
\end{algorithm}

\clearpage
\section{Multi-Task Learning}
\label{sec:exp_multi}
\begin{figure*}[h!]
\captionsetup[subfigure]{position=above,labelformat=empty,margin={1em,0em},skip=-.1em}
\centering
\begin{subfigure}[t]{.35\textwidth}
\centering
\caption{Average over tasks}
\includegraphics[width=\textwidth]{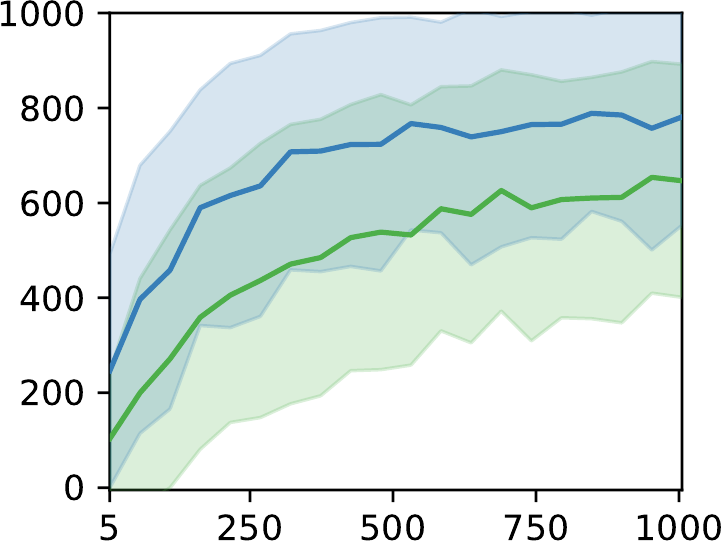}
\end{subfigure}\hfill%
\caption{We compare a single \method agent trained on all tasks to individual \method agents. The plot shows test performance over the number of episodes collected for each task. The single agent learns to solve all the tasks while learning more slowly compared to the individual agents. The lines show mean and one standard deviation over 6 tasks, 5 seeds, and 10 trajectories.}
\label{fig:score_multi_avg}
\end{figure*}

\begin{figure*}[h!]
\providecommand{\width}{}
\renewcommand{\width}{0.32\textwidth}
\providecommand{\path}{}
\renewcommand{\path}{score_multi}
\captionsetup[subfigure]{position=above,labelformat=empty,margin={1em,0em},skip=-.1em}
\centering\hfil\hfil%
\begin{subfigure}[t]{\width}
\centering
\caption{Cartpole Swing Up}
\includegraphics[width=\textwidth]{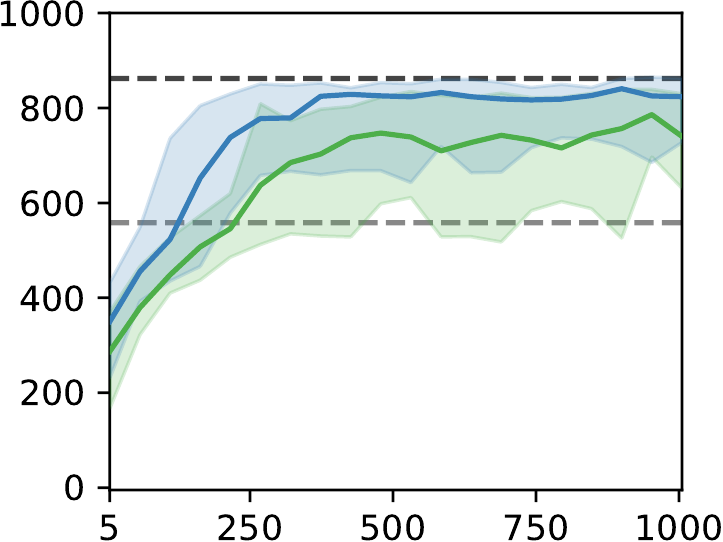}
\end{subfigure}\hfil%
\begin{subfigure}[t]{\width}
\centering
\caption{Reacher Easy}
\includegraphics[width=\textwidth]{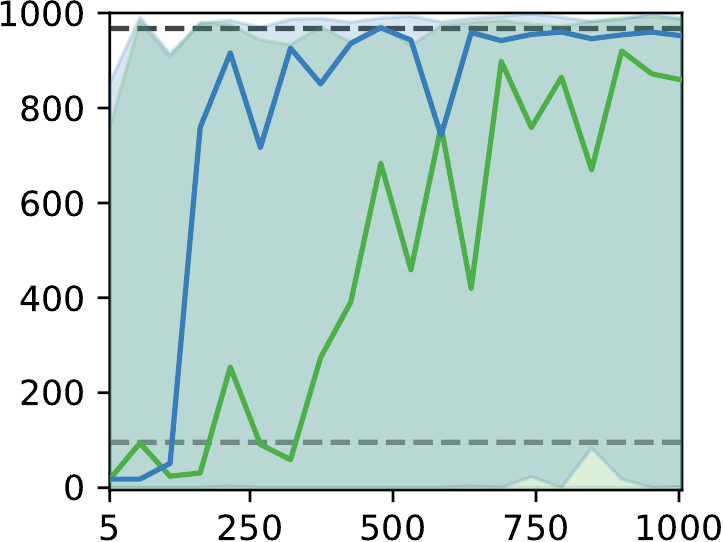}
\end{subfigure}\hfil%
\begin{subfigure}[t]{\width}
\centering
\caption{Cheetah Run}
\includegraphics[width=\textwidth]{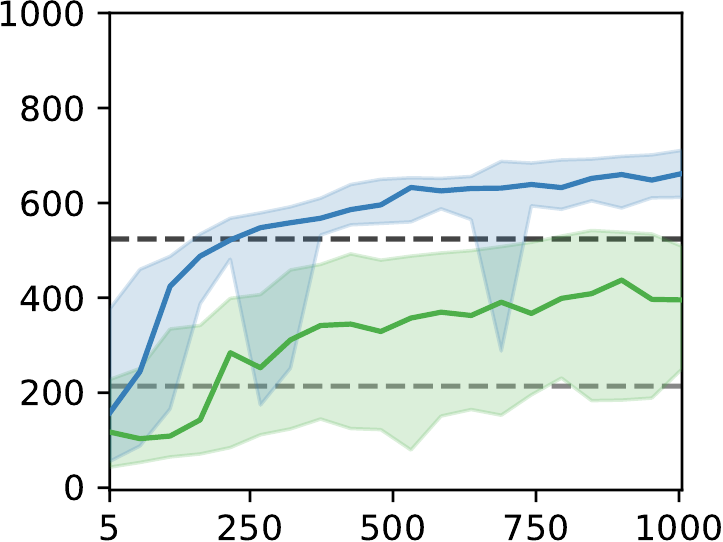}
\end{subfigure}\hfil\hfil\hfil \\
\vspace{.7em}\hfil\hfil%
\begin{subfigure}[t]{\width}
\centering
\caption{Finger Spin}
\includegraphics[width=\textwidth]{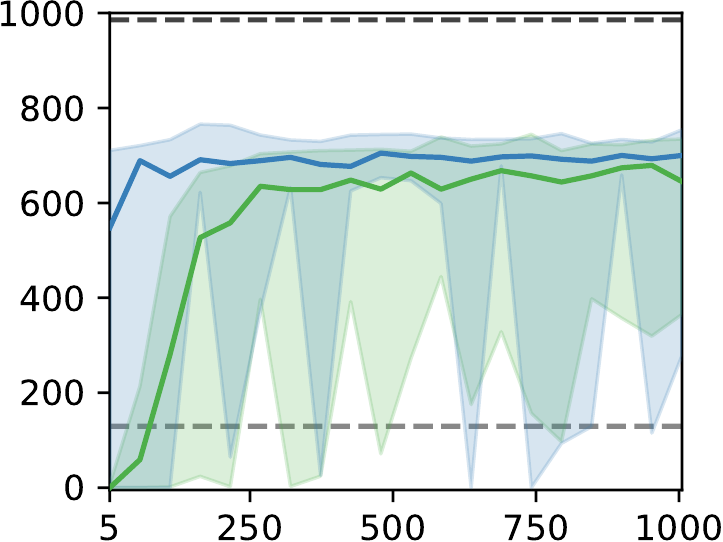}
\end{subfigure}\hfil%
\begin{subfigure}[t]{\width}
\centering
\caption{Cup Catch}
\includegraphics[width=\textwidth]{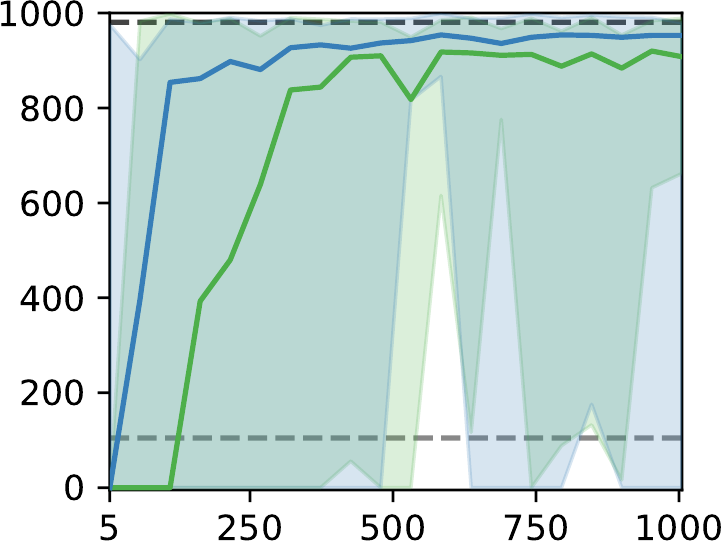}
\end{subfigure}\hfil%
\begin{subfigure}[t]{\width}
\centering
\caption{Walker Walk}
\includegraphics[width=\textwidth]{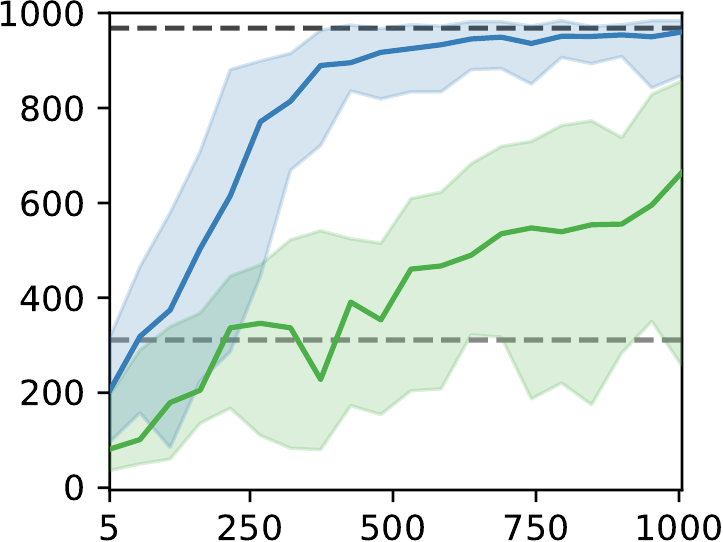}
\end{subfigure}\hfil\hfil\hfil \\
\vspace{1em}\hspace{1em}
\begin{subfigure}[t]{.52\textwidth}
\centering
\includegraphics[width=\textwidth]{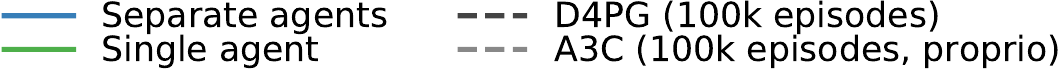}
\end{subfigure}\hfil%
\caption{Per-task performance of a single PlaNet agent trained on the six tasks. Plots show test performance over the number of episodes collected per task. The agent is not told which task it is solving and it needs to infer this from the image observations. The agent learns to distinguish the tasks and solve them with just a moderate slowdown in learning. The lines show medians and the areas show percentiles 5 to 95 over 4 seeds and 10 trajectories.}
\label{fig:score_multi}
\end{figure*}

\clearpage
\section{Latent Overshooting}
\label{sec:exp_overshooting}
\begin{figure*}[h!]
\providecommand{\width}{}
\renewcommand{\width}{0.32\textwidth}
\providecommand{\path}{}
\renewcommand{\path}{score_overshooting}
\captionsetup[subfigure]{position=above,labelformat=empty,margin={1em,0em},skip=-.1em}
\centering\hfil\hfil%
\begin{subfigure}[t]{\width}
\centering
\caption{Cartpole Swing Up}
\includegraphics[width=\textwidth]{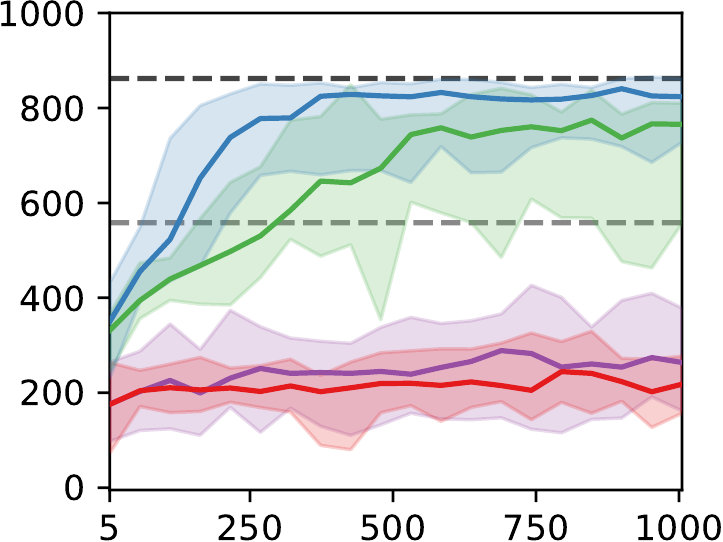}
\end{subfigure}\hfil%
\begin{subfigure}[t]{\width}
\centering
\caption{Reacher Easy}
\includegraphics[width=\textwidth]{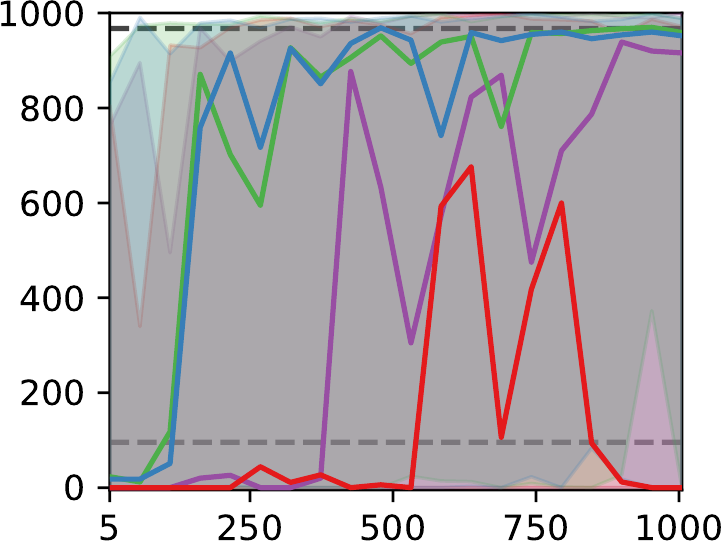}
\end{subfigure}\hfil%
\begin{subfigure}[t]{\width}
\centering
\caption{Cheetah Run}
\includegraphics[width=\textwidth]{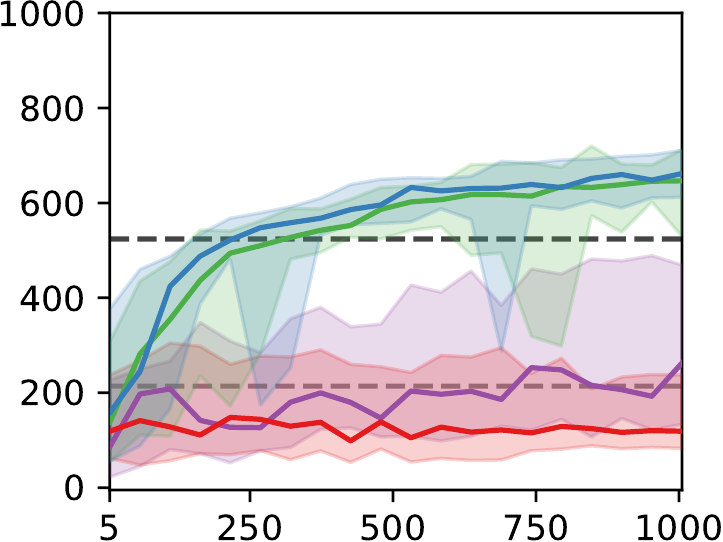}
\end{subfigure}\hfil\hfil\hfil \\
\vspace{.7em}\hfil\hfil%
\begin{subfigure}[t]{\width}
\centering
\caption{Finger Spin}
\includegraphics[width=\textwidth]{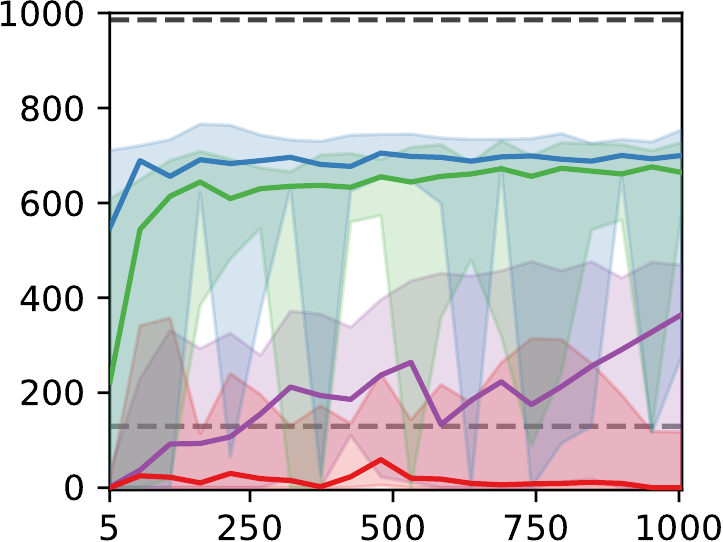}
\end{subfigure}\hfil%
\begin{subfigure}[t]{\width}
\centering
\caption{Cup Catch}
\includegraphics[width=\textwidth]{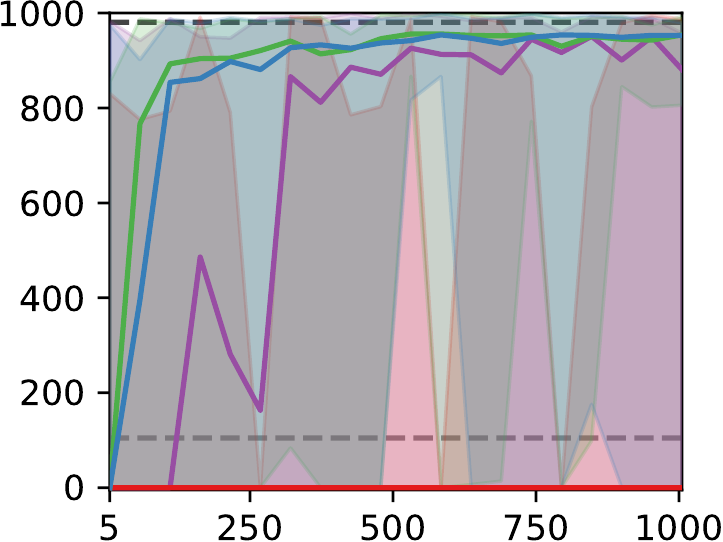}
\end{subfigure}\hfil%
\begin{subfigure}[t]{\width}
\centering
\caption{Walker Walk}
\includegraphics[width=\textwidth]{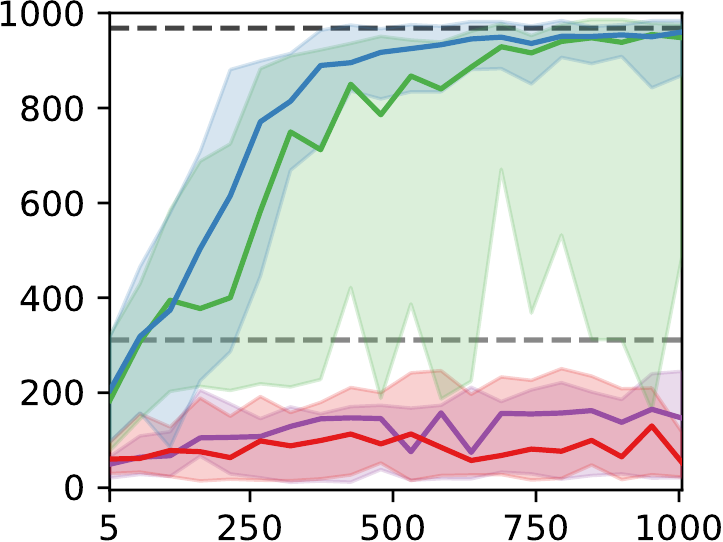}
\end{subfigure}\hfil\hfil\hfil \\
\vspace{1em}\hspace{1em}
\begin{subfigure}[t]{.82\textwidth}
\centering
\includegraphics[width=\textwidth]{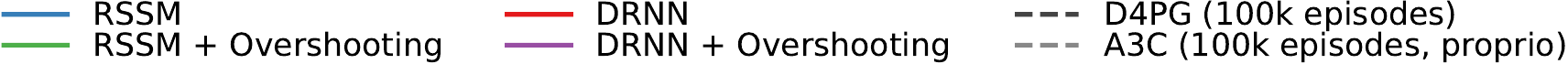}
\end{subfigure}\hfil%
\caption{We compare the standard variational objective with latent overshooting on our proposed RSSM and another model called DRNN that uses two RNNs as encoder and decoder with a stochastic state sequence in between. Latent overshooting can substantially improve the performance of the DRNN and other models we have experimented with (not shown), but slightly reduces performance of our RSSM. The lines show medians and the areas show percentiles 5 to 95 over 5 seeds and 10 trajectories.}
\label{fig:score_overshooting}
\end{figure*}

\clearpage
\section{Activation Function}
\label{sec:exp_activation}
\begin{figure*}[h!]
\providecommand{\width}{}
\renewcommand{\width}{0.32\textwidth}
\providecommand{\path}{}
\renewcommand{\path}{score_activation}
\captionsetup[subfigure]{position=above,labelformat=empty,margin={1em,0em},skip=-.1em}
\centering\hfil\hfil%
\begin{subfigure}[t]{\width}
\centering
\caption{Cartpole Swing Up}
\includegraphics[width=\textwidth]{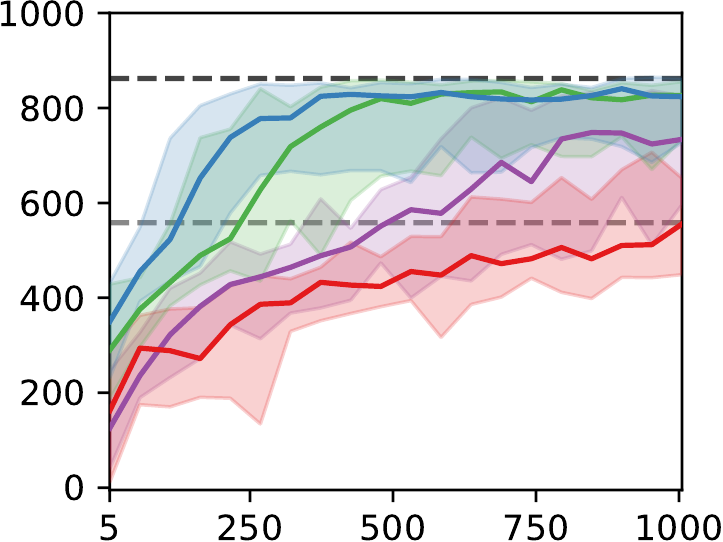}
\end{subfigure}\hfil%
\begin{subfigure}[t]{\width}
\centering
\caption{Reacher Easy}
\includegraphics[width=\textwidth]{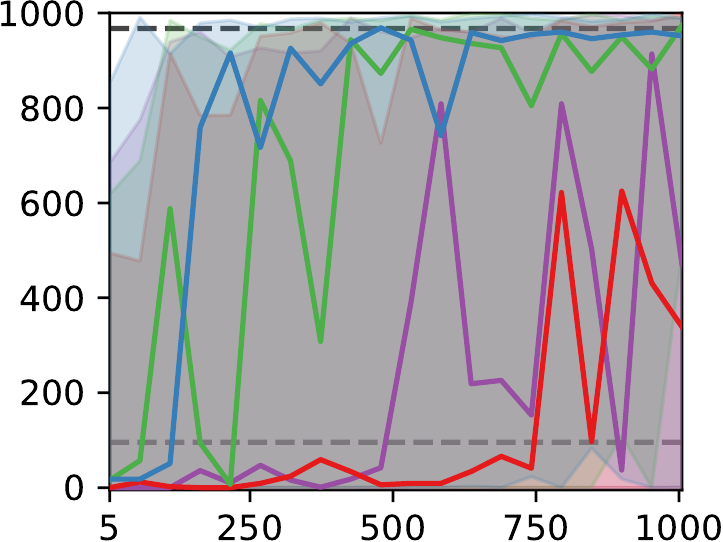}
\end{subfigure}\hfil%
\begin{subfigure}[t]{\width}
\centering
\caption{Cheetah Run}
\includegraphics[width=\textwidth]{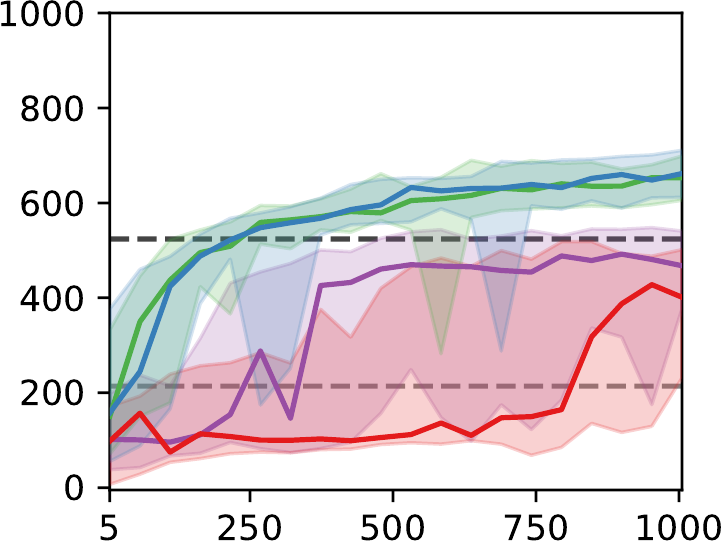}
\end{subfigure}\hfil\hfil\hfil \\
\vspace{.7em}\hfil\hfil%
\begin{subfigure}[t]{\width}
\centering
\caption{Finger Spin}
\includegraphics[width=\textwidth]{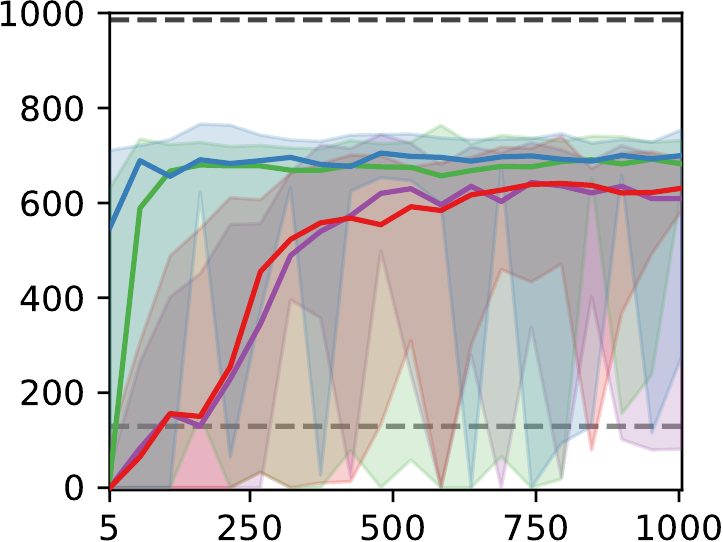}
\end{subfigure}\hfil%
\begin{subfigure}[t]{\width}
\centering
\caption{Cup Catch}
\includegraphics[width=\textwidth]{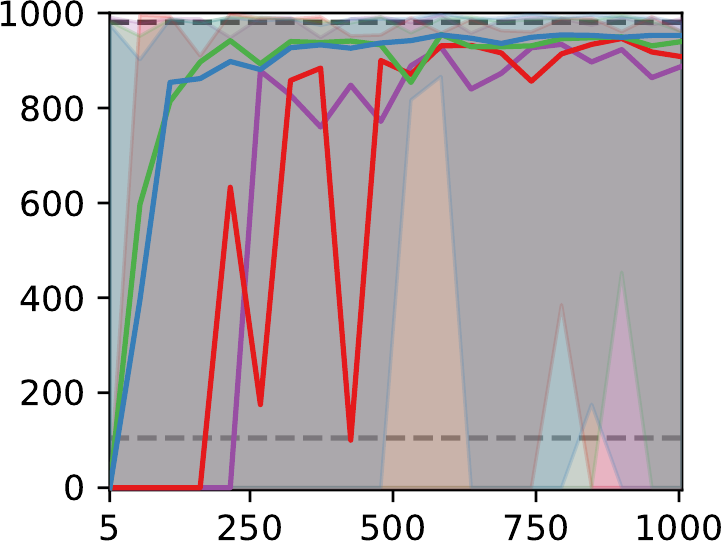}
\end{subfigure}\hfil%
\begin{subfigure}[t]{\width}
\centering
\caption{Walker Walk}
\includegraphics[width=\textwidth]{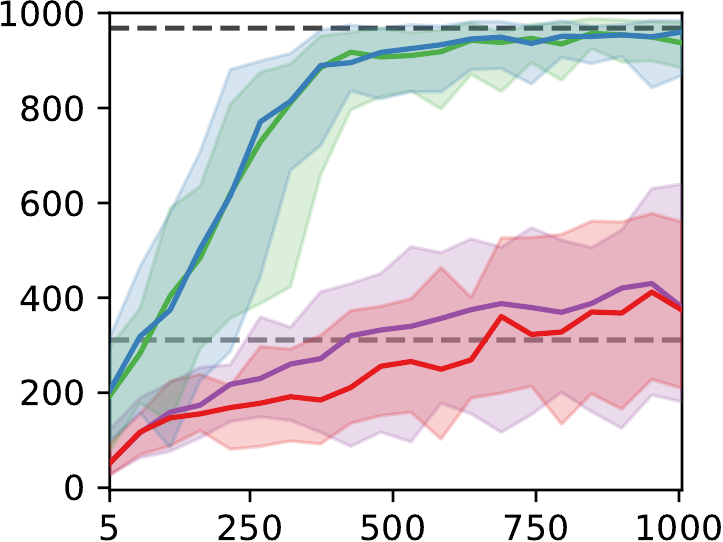}
\end{subfigure}\hfil\hfil\hfil \\
\vspace{1em}\hspace{1em}
\begin{subfigure}[t]{.72\textwidth}
\centering
\includegraphics[width=\textwidth]{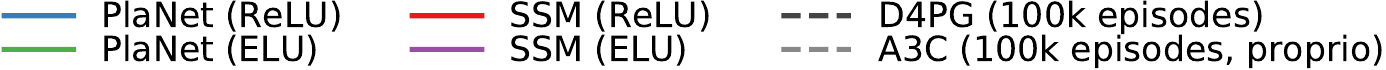}
\end{subfigure}\hfil%
\caption{Comparison of hard ReLU \citep{nair2010relu} and smooth ELU \citep{clevert2015elu} activation functions. We find that smooth activations help improve performance of the purely stochastic model (and the purely deterministic model; not shown) while our proposed RSSM is robust to the choice of activation function. The lines show medians and the areas show percentiles 5 to 95 over 5 seeds and 10 trajectories.}
\label{fig:score_activation}
\end{figure*}

\clearpage
\section{Bound Derivations}
\label{sec:derivs}

\paragraph{One-step predictive distribution}

The variational bound for latent dynamics models $\p(o_{1:T},s_{1:T}|a_{1:T})=\prod_t\p(s_t|s_{t-1},a_{t-1})\p(o_t|s_t)$ and a variational posterior $\q(s_{1:T}|o_{1:T},a_{1:T})=\prod_t\q(s_t|o_{\leq t},a_{<t})$ follows from importance weighting and Jensen's inequality as shown,
\begin{equation}
\begin{aligned}
\ln\p(o_{1:T}|a_{1:T})
&\triangleq\ln\E[\bigg]{\p(s_{1:T}|a_{1:T})}{\prod_{t=1}^T \p(o_t|s_t)} \\
&=\ln\E[\bigg]{\q(s_{1:T}|o_{1:T},a_{1:T})}{\prod_{t=1}^T \p(o_t|s_t)\p(s_t|s_{t-1},a_{t-1})/\q(s_t|o_{\leq t},a_{<t})} \\
&\geq\E[\bigg]{\q(s_{1:T}|o_{1:T},a_{1:T})}{\sum_{t=1}^T \ln\p(o_t|s_t)+\ln\p(s_t|s_{t-1},a_{t-1})-\ln\q(s_t|o_{\leq t},a_{<t})} \\
&=\sum_{t=1}^T \Big(
  \describe{\Ebelow{\q(s_t|o_{\leq t},a_{<t})}{\ln\p(o_t|s_t)}}{reconstruction}
  -\describe{\Ebelow[\big]{\q(s_{t-1|o_{\leq t-1},a_{<a-1}})}{\KL{\q(s_t|o_{\leq t},a_{<t})}{\p(s_t|s_{t-1},a_{t-1})}}}{complexity} \Big).
\end{aligned}
\label{eq:elbo_deriv}
\end{equation}%

\paragraph{Multi-step predictive distribution}

The variational bound on the $d$-step predictive distribution $\pr[p_d](o_{1:T},s_{1:T}|a_{1:T})=\prod_t\p(s_t|s_{t-d},a_{t-1})\p(o_t|s_t)$ and a variational posterior $\q(s_{1:T}|o_{1:T},a_{1:T})=\prod_t\q(s_t|o_{\leq t},a_{<t})$ follows analogously. The second bound comes from moving the log inside the multi-step priors, which satisfy the recursion $\p(s_t|s_{t-d},a_{t-d-1:t-1})=\E{\p(s_{t-1}|s_{t-d},a_{t-d-1:t-2})}{\p(s_t|s_{t-1},a_{t-1})}$.
\begin{equation}
\begin{aligned}
\ln\pr[p_d](o_{1:T}|a_{1:T})
&\triangleq\ln\E[\bigg]{\pr[p_d](s_{1:T}|a_{1:T})}{\prod_{t=1}^T\p(o_t|s_t)} \\
&=\ln\E[\bigg]{\q(s_{1:T}|o_{1:T},a_{1:T})}{\prod_{t=1}^T \p(o_t|s_t)\p(s_t|s_{t-d},a_{t-d-1:t-1})/\q(s_t|o_{\leq t},a_{<t})} \\
&\geq\E[\bigg]{\q(s_{1:T}|o_{1:T},a_{1:T})}{\sum_{t=1}^T \ln\p(o_t|s_t)+\ln\p(s_t|s_{t-d},a_{t-d-1:t-1})-\ln\q(s_t|o_{\leq t},a_{<t})} \\
&\geq\E[\bigg]{\q(s_{1:T}|o_{1:T},a_{1:T})}{\sum_{t=1}^T \ln\p(o_t|s_t)+\Ebelow{\p(s_{t-1}|s_{t-d},a_{t-d-1:t-2})}{\ln\p(s_t|s_{t-1},a_{t-1})}-\ln\q(s_t|o_{\leq t},a_{<t})} \\
&=\sum_{t=1}^T \Big(
  \describe{\E{\q(s_t|o_{\leq t},a_{<t})}{\ln\p(o_t|s_t)}}{reconstruction}
  -\describe{\Ebelow[\big]{\p(s_{t-1}|s_{t-d},a_{t-d-1:t-2})\q(s_{t-d}|o_{\leq t-d},a_{<t-d})}{\KL{\q(s_t|o_{\leq t},a_{<t})}{\p(s_t|s_{t-1},a_{t-1})}}}{multi-step prediction} \Big).
\end{aligned}
\label{eq:dstep_deriv}
\end{equation}%
Since all expectations are on the outside of the objective, we can easily obtain an unbiased estimator of this bound by changing expectations to sample averages.

\paragraph{Relation between one-step and multi-step predictive distributions}

We conjecture that the multi-step predictive distribution $\pr[p_d](o_{1:T})$ lower bounds the one-step predictive distribution $\p(o_{1:T})$ of the same latent sequence model model in expectation over the data set. Since the latent state sequence is Markovian, for $d\geq 1$ we have the data processing inequality
\begin{gather}
\begin{aligned}
\MI{s_t;s_{t-d}} &\leq \MI{s_t;s_{t-1}} \\
\H{s_t}-\H{s_t|s_{t-d}} &\leq \H{s_t}-\H{s_t|s_{t-1}} \\
\E{}{\ln\p(s_t|s_{t-d})} &\leq \E{}{\ln\p(s_t|s_{t-1})} \\
\E{}{\ln\pr[p_d](o_{1:T})} &\leq \E{}{\ln\p(o_{1:T})}.
\end{aligned}
\label{eq:dataproc_deriv}
\end{gather}
Therefore, any bound on the multi-step predictive distribution, including \cref{eq:dstep_deriv} and \cref{eq:latov}, is also a bound on the one-step predictive distribution.

\clearpage
\section{Additional Related Work}
\label{sec:more_related_work}

\paragraph{Planning in state space}

When low-dimensional states of the environment are available to the agent, it is possible to learn the dynamics directly in state space. In the regime of control tasks with only a few state variables, such as the cart pole and mountain car tasks, PILCO \citep{deisenroth2011pilco} achieves remarkable sample efficiency using Gaussian processes to model the dynamics. Similar approaches using neural networks dynamics models can solve two-link balancing problems \citep{gal2016deeppilco,higuera2018synthesizing} and implement planning via gradients \citep{henaff2018planbybackprop}. \citet{chua2018pets} use ensembles of neural networks, scaling up to the cheetah running task. The limitation of these methods is that they access the low-dimensional Markovian state of the underlying system and sometimes the reward function. \citet{amos2018awareness} train a deterministic model using overshooting in observation space for active exploration with a robotics hand. We move beyond low-dimensional state representations and use a latent dynamics model to solve control tasks from images.

\paragraph{Hybrid agents}

The challenges of model-based RL have motivated the research community to develop hybrid agents that accelerate policy learning by training on imagined experience \citep{kalweit2017blending,nagabandi2017mbmf,kurutach2018modeltrpo,buckman2018steve,ha2018worldmodels}, improving feature representations \citep{wayne2018merlin,igl2018dvrl}, or leveraging the information content of the model directly \citep{weber2017i2a}. \citet{srinivas2018upn} learn a policy network with integrated planning computation using reinforcement learning and without prediction loss, yet require expert demonstrations for training.

\paragraph{Multi-step predictions}

Training sequence models on multi-step predictions has been explored for several years. Scheduled sampling \citep{bengio2015scheduled} changes the rollout distance of the sequence model over the course of training. Hallucinated replay \citep{talvitie2014hallucinated} mixes predictions into the data set to indirectly train multi-step predictions. \citet{venkatraman2015dad} take an imitation learning approach. Recently, \citet{amos2018awareness} train a dynamics model on all multi-step predictions at once. We generalize this idea to latent sequence models trained via variational inference.

\paragraph{Latent sequence models}

Classic work has explored models for non-Markovian observation sequences, including recurrent neural networks (RNNs) with deterministic hidden state and probabilistic state-space models (SSMs). The ideas behind variational autoencoders \citep{kingma2013vae,rezende2014vae} have enabled non-linear SSMs that are trained via variational inference \citep{krishnan2015deepkalman}. The VRNN \citep{chung2015vrnn} combines RNNs and SSMs and is trained via variational inference. In contrast to our RSSM, it feeds generated observations back into the model which makes forward predictions expensive. \citet{karl2016dvbf} address mode collapse to a single future by restricting the transition function, \citep{moerland2017learning} focus on multi-modal transitions, and \citet{doerr2018prssm} stabilize training of purely stochastic models. \citet{buesing2018dssm} propose a model similar to ours but use in a hybrid agent instead for explicit planning.

\paragraph{Video prediction}

Video prediction is an active area of research in deep learning. \citet{oh2015atari} and \citet{chiappa2017recurrent} achieve visually plausible predictions on Atari games using deterministic models. \citet{kalchbrenner2016vpn} introduce an autoregressive video prediction model using gated CNNs and LSTMs. Recent approaches introduce stochasticity to the model to capture multiple futures \citep{babaeizadeh2017sv2p,denton2018stochastic}. To obtain realistic predictions, \citet{mathieu2015deep} and \citet{vondrick2016generating} use adversarial losses. In simulated environments, \citet{gemici2017temporalmemory} augment dynamics models with an external memory to remember long-time contexts. \citet{van2017vq} propose a variational model that avoids sampling using a nearest neighbor look-up, yielding high fidelity image predictions. These models are complimentary to our approach.

\clearpage
\section{Video Predictions}
\label{sec:prediction}
\begin{figure*}[h]
\providecommand{\width}{}
\renewcommand{\width}{0.62\textwidth}
\centering
\begin{subfigure}[t]{\width}
\rot{\hspace{4.6em}\method}
\hspace{.5em}
\includegraphics[width=\textwidth]{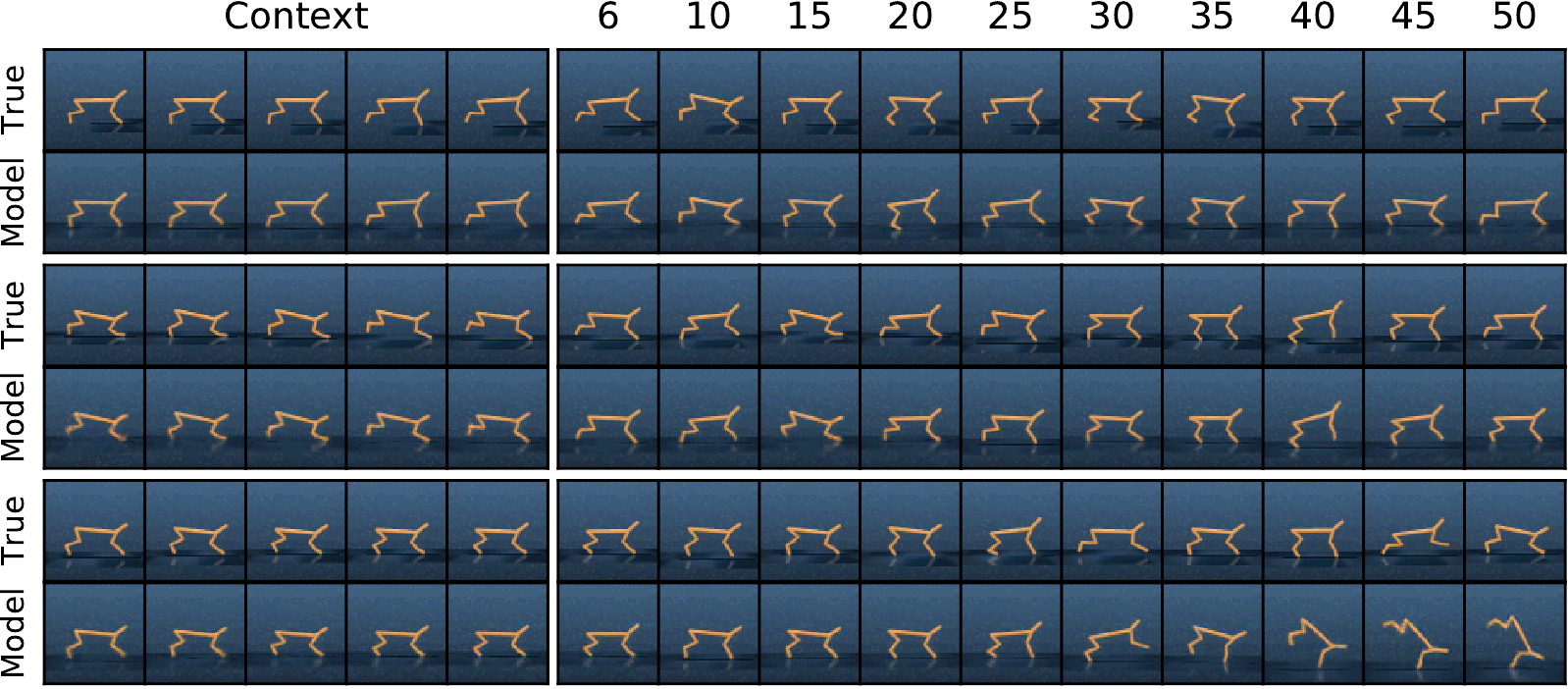}
\end{subfigure} \\
\vspace{2ex}
\begin{subfigure}[t]{\width}
\rot{\hspace{1.3em}\method + Overshooting}
\hspace{.5em}
\includegraphics[width=\textwidth]{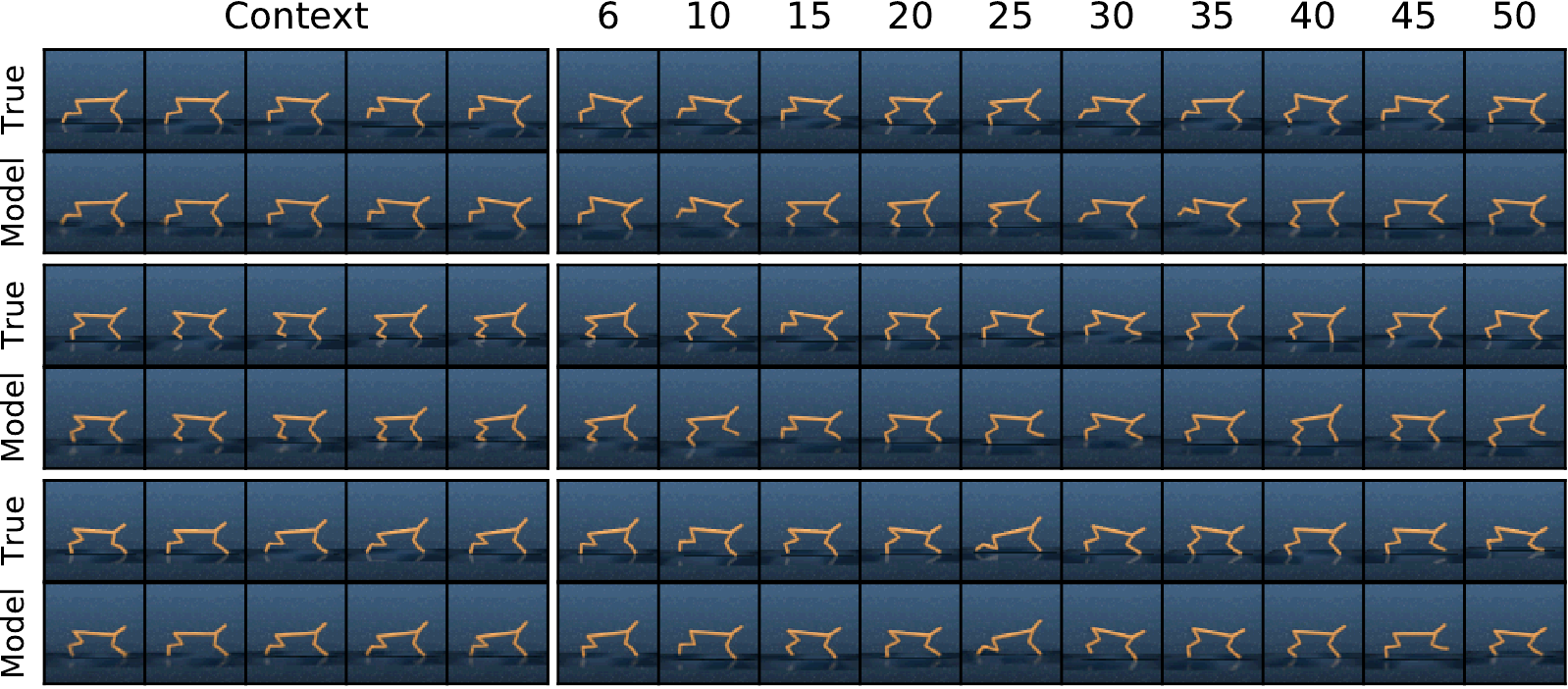}
\end{subfigure} \\
\vspace{2ex}
\begin{subfigure}[t]{\width}
\rot{\hspace{1.8em}Deterministic (GRU)}
\hspace{.5em}
\includegraphics[width=\textwidth]{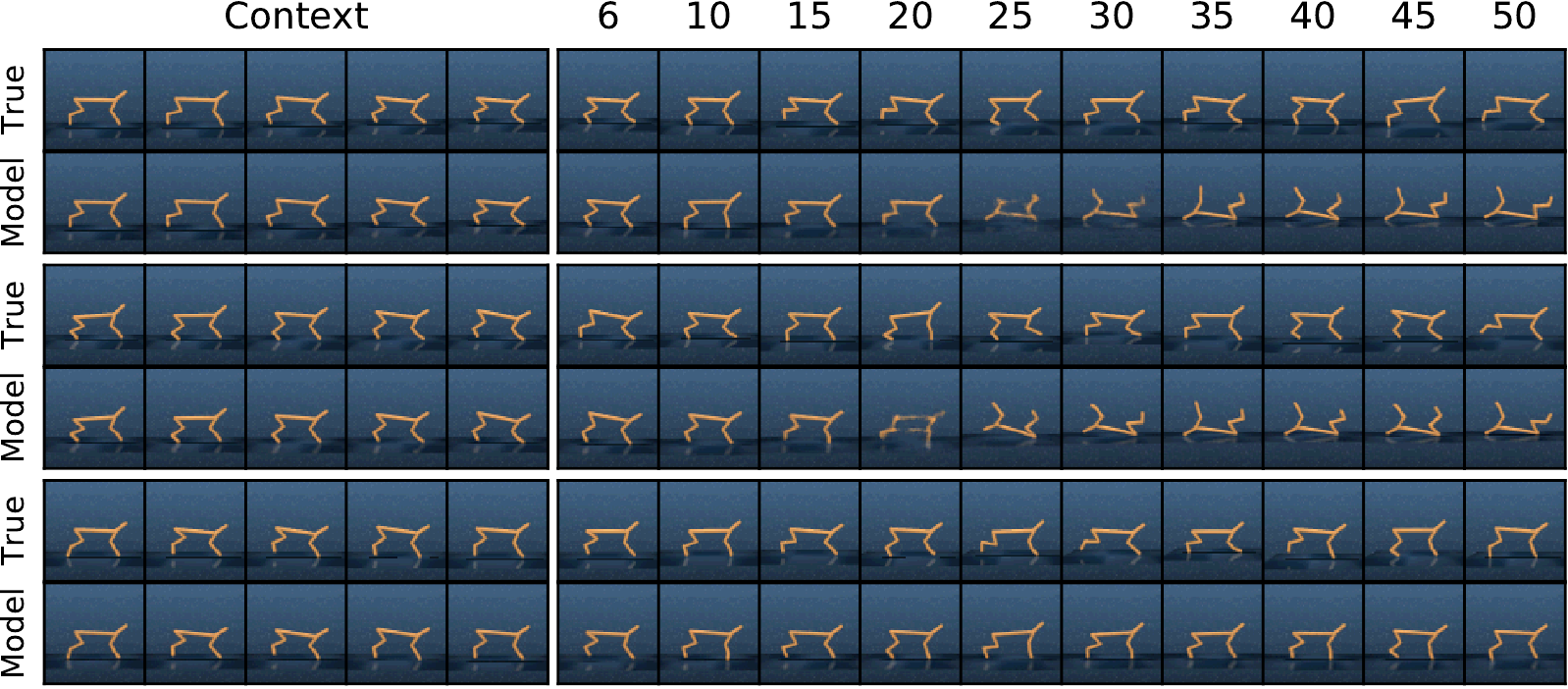}
\end{subfigure} \\
\vspace{2ex}
\begin{subfigure}[t]{\width}
\rot{\hspace{2.5em}Stochastic (SSM)}
\hspace{.5em}
\includegraphics[width=\textwidth]{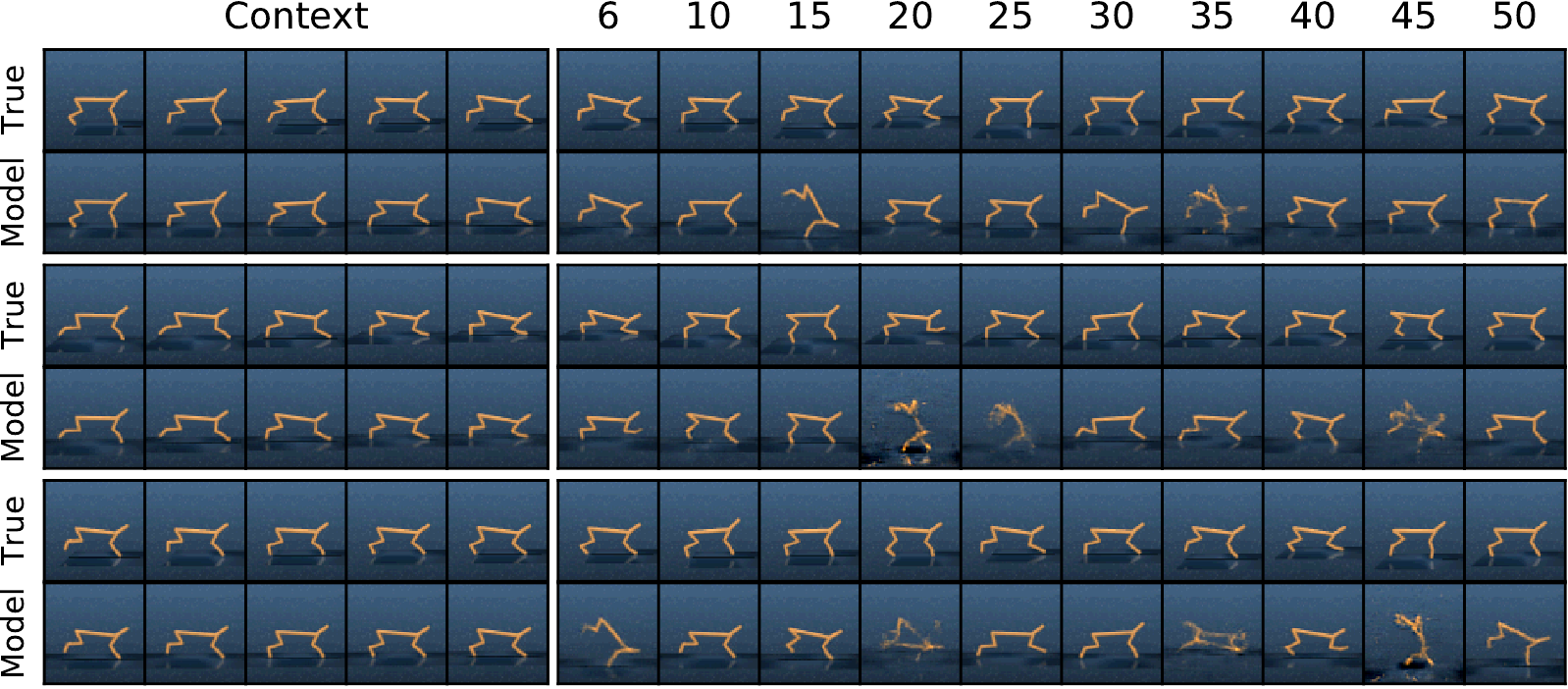}
\end{subfigure} \\
\caption{Open-loop video predictions for test episodes. The \mbox{columns 1--5} show reconstructed context frames and the remaining images are generated open-loop. Our RSSM achieves pixel-accurate predictions for 50 steps into the future in the cheetah environment. We randomly selected action sequences from test episodes collected with action noise alongside the training episodes.}
\label{fig:prediction}
\vspace{-0.2in}
\end{figure*}

\clearpage
\section{State Diagnostics}

\begin{figure}[h]
\centering
\includegraphics[width=\textwidth]{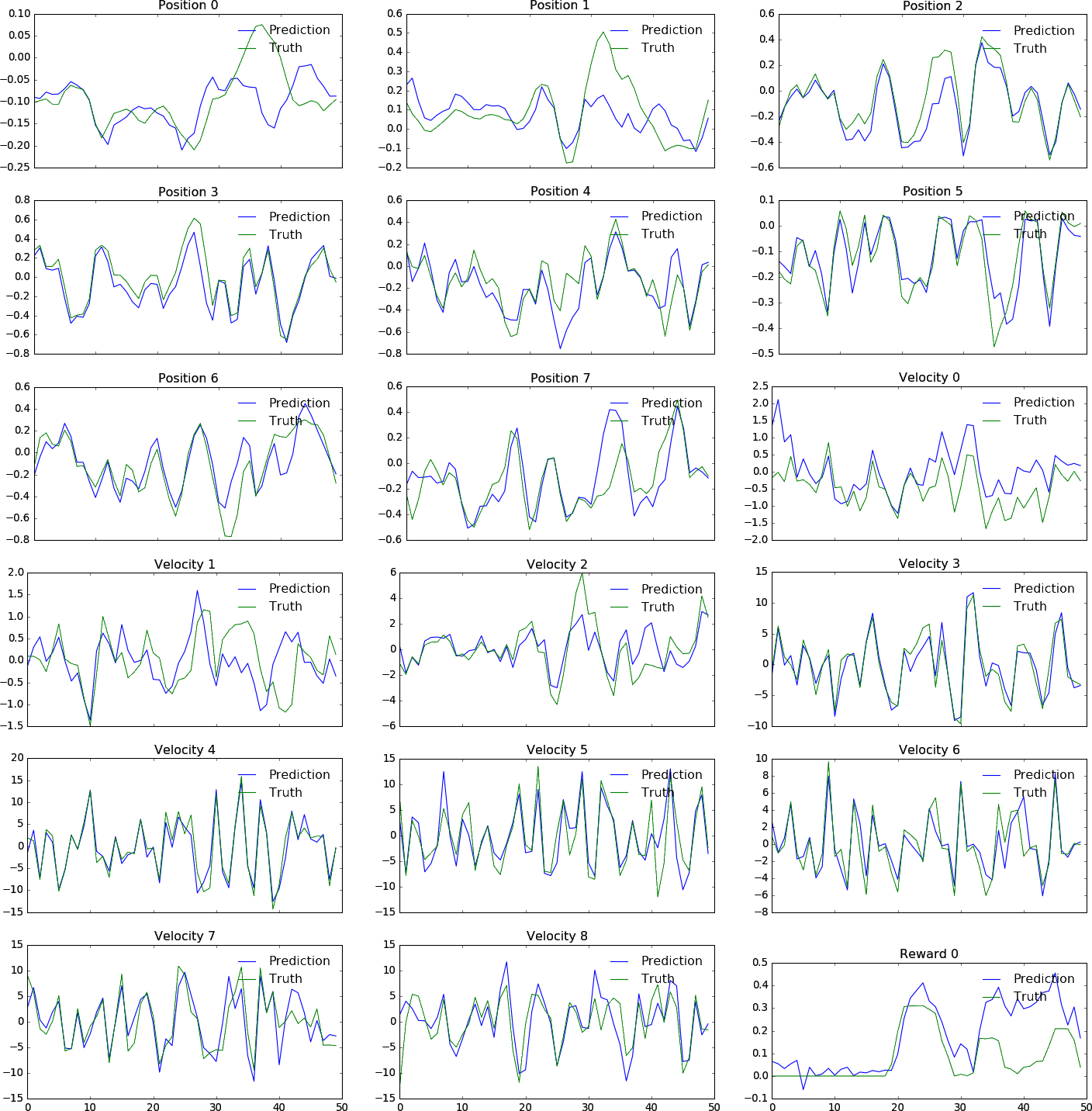}
\caption{Open-loop state diagnostics. We freeze the dynamics model of a \method agent and learn small neural networks to predict the true positions, velocities, and reward of the simulator. The open-loop predictions of these quantities show that most information about the underlying system is present in the learned latent space and can be accurately predicted forward further than the planning horizons used in this work.}
\label{fig:diagnostics}
\end{figure}

\clearpage
\section{Planning Parameters}

\begin{figure}[h!]
\centering
\includegraphics[width=\textwidth]{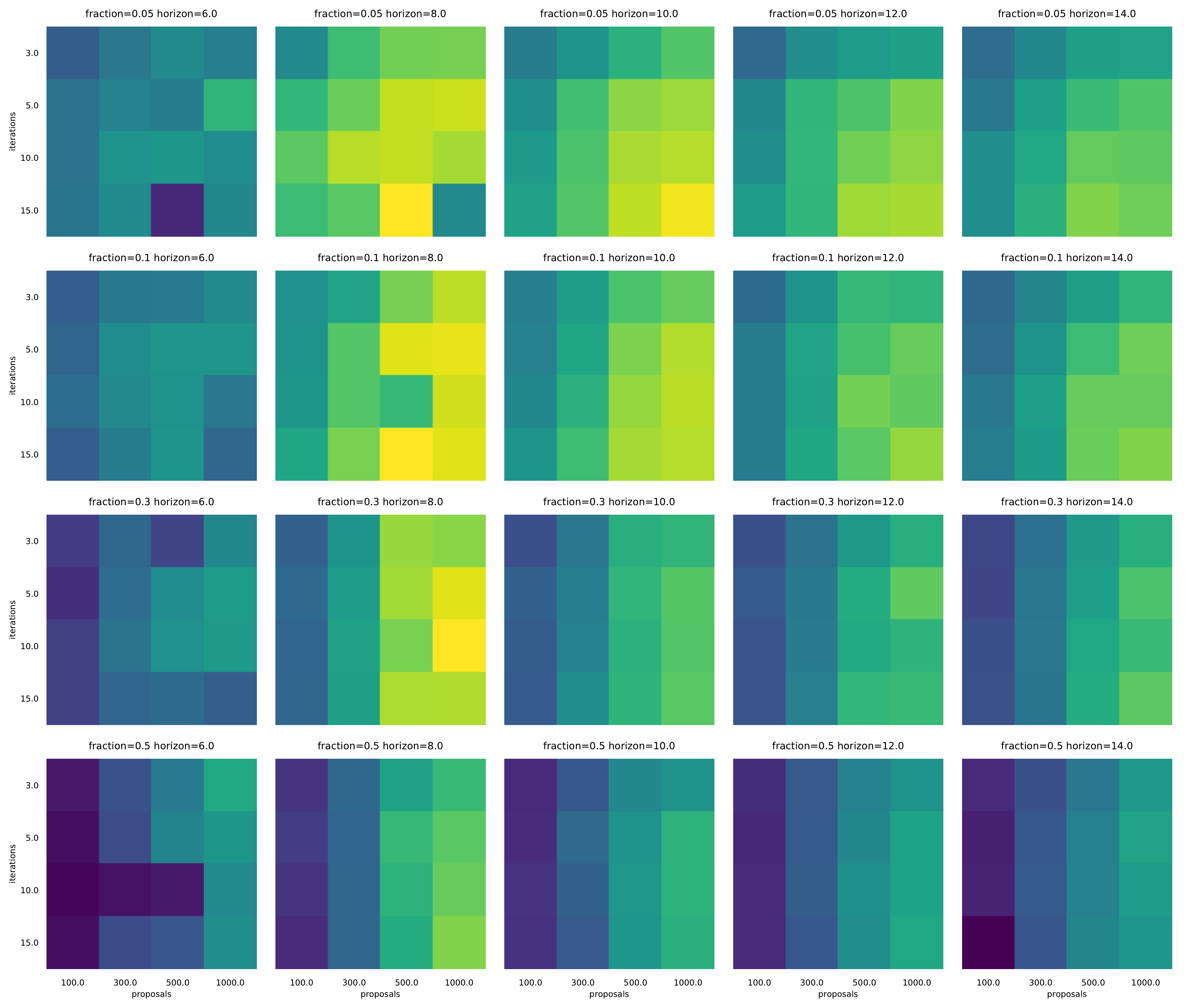}
\caption{Planning performance on the cheetah running task with the true simulator using different planner settings. Performance ranges from 132 (blue) to 837 (yellow). Evaluating more action sequences, optimizing for more iterations, and re-fitting to fewer of the best proposals tend to improve performance. A planning horizon length of 6 is not sufficient and results in poor performance. Much longer planning horizons hurt performance because of the increased search space. For this environment, best planning horizon length is near 8 steps.}
\end{figure}

\end{document}